\pdfoutput=1

\documentclass[11pt]{article}

\usepackage{acl}

\usepackage{times}
\usepackage{latexsym}

\usepackage[T1]{fontenc}

\usepackage[utf8]{inputenc}

\usepackage{microtype}

\usepackage{inconsolata}
\usepackage{microtype}
\usepackage{graphicx}
\usepackage{import}
\usepackage{layout}
\usepackage{tabularx, makecell}
\usepackage{booktabs}
\usepackage{mathrsfs}
\usepackage{amssymb} 
\usepackage{url}
\usepackage{graphicx}
\usepackage{xspace,paralist}
\usepackage{times,latexsym}
\usepackage{amsmath}
\usepackage{appendix}
\usepackage{comment}
\usepackage{adjustbox}
\usepackage{enumitem}
\usepackage{makecell}
\usepackage{multirow}
\usepackage{xcolor}
\usepackage{cleveref}
\usepackage{todonotes}
\usepackage{longtable,supertabular}
\usepackage{amssymb}
\usepackage{pifont}

\newcommand{\tabref}[2][]{Table#1~\ref{#2}\xspace}

\newcommand{\dataset}[1]{\text{#1}\xspace}
\newcommand{\arxiv}{\dataset{arXiv}}
\newcommand{\wikipedia}{\dataset{Wikipedia}}
\newcommand{\wikihow}{\dataset{WikiHow}}
\newcommand{\peerread}{\dataset{PeerRead}}
\newcommand{\reddit}{\dataset{Reddit}}
\newcommand{\outfox}{\dataset{OUTFOX}}

\newcommand{\model}[1]{\text{#1}\xspace}

\newcommand{\chatgpt}{\model{ChatGPT}}
\newcommand{\gptfour}{\model{GPT-4}}

\newcommand{\davinci}{\model{davinci-003}}

\newcommand{\cohere}{\model{Cohere}}
\newcommand{\dolly}{\model{Dolly-v2}}
\newcommand{\bloomz}{\model{BLOOMz}}

\newcommand{\llamatwo}{\model{LLaMA-2}}

\title{SemEval-2024 Task 8: Multidomain, Multimodel and Multilingual\\ Machine-Generated Text Detection}

\author{
Yuxia Wang,\textsuperscript{\textdagger} 
Jonibek Mansurov,\textsuperscript{\textdagger}
Petar Ivanov,\textsuperscript{\textdagger}
Jinyan Su,\textsuperscript{\textdagger}
Artem Shelmanov,\textsuperscript{\textdagger}\\ 
{\bf 
Akim Tsvigun,\textsuperscript{\textdagger}
Osama Mohammed Afzal,\textsuperscript{\textdagger}
Tarek Mahmoud,\textsuperscript{\textdagger}
}\\
{\bf  
Giovanni Puccetti,\textsuperscript{\S}
Thomas Arnold,\textsuperscript{\textparagraph}
Chenxi Whitehouse,\textsuperscript{\textasteriskcentered}
}\\
{\bf Alham Fikri Aji,\textsuperscript{\textdagger} 
Nizar Habash,\textsuperscript{\textdagger}\textsuperscript{\textdaggerdbl} 
Iryna Gurevych,\textsuperscript{\textdagger} 
Preslav Nakov\textsuperscript{\textdagger}
}\\
  \textsuperscript{\textdagger}MBZUAI, UAE 
  \textsuperscript{\textparagraph}TU Darmstadt, Germany  
  \textsuperscript{\textasteriskcentered}University of Cambridge, UK \\
  \textsuperscript{\S}Institute of Information Science and Technology, Italy  
  \textsuperscript{\textdaggerdbl}New York University Abu Dhabi, UAE \\
  \texttt{\{yuxia.wang, jonibek.mansurov, preslav.nakov\}@mbzuai.ac.ae} \\
}

\begin{document}
\maketitle
\begin{abstract}
We present the results and the main findings of SemEval-2024 Task 8: Multigenerator, Multidomain, and Multilingual Machine-Generated Text Detection. The task featured three subtasks. Subtask A is a binary classification task determining whether a text is written by a human or generated by a machine. This subtask has two tracks: a monolingual track focused solely on English texts and a multilingual track. Subtask B is to detect the exact source of a text, discerning whether it is written by a human or generated by a specific LLM. Subtask C aims to identify the changing point within a text, at which the authorship transitions from human to machine. 
The task attracted a large number of participants: subtask A monolingual (126), subtask A multilingual (59), subtask B (70), and subtask C (30). 
In this paper, we present the task, analyze the results, and discuss the system submissions and the methods they used. For all subtasks, the best systems used LLMs.

\end{abstract}

\section{Introduction}


The proliferation of Large Language Models (LLMs) has led to a significant increase in the volume of machine-generated text (MGT) across a wide range of domains. This rise has sparked concerns regarding the potential for misuse in fields such as journalism, education, academia, etc \cite{uchendu2023attribution,crothers2023machine}. Moreover, it poses challenges to maintaining information integrity and ensuring accurate information dissemination. As such, the ability to accurately distinguish between human-written content and machine-generated content has become paramount for identifying potential misuse \cite{jawahar2020automatic,stiff2022detecting,macko-etal-2023-multitude}.

In response to these challenges, we are introducing a shared task that focuses on the detection of machine-generated text across multiple generators, domains, and languages. We are providing large-scale evaluation datasets for \textbf{three subtasks} with the primary goals of fostering extensive research in MGT detection, advancing the development of automated systems for detecting MGT, and reducing instances of misuse: 
%
 
    \paragraph{Subtask A: Human vs. Machine Classification.} The goal of this subtask is to accurately classify a text as either produced by a human or generated by a machine. This is the basic, but one of the most common use-cases of MGT detection systems for preventing the misuse of LLMs. This task is divided into two tracks:
        (i) The \emph{monolingual track}, which focuses solely on English texts; and
        (ii) The \emph{multilingual track}, which involves texts in a variety of languages, thereby expanding the diversity and complexity beyond existing benchmarks.
 
    \paragraph{Subtask B: Multi-Way Generator Detection.} This task aims to pinpoint the exact source of a text, i.e., determine whether it originated from a human or a specific LLM (GPT-3, GPT-3.5, GPT-4, Cohere, DALL-E, or BLOOMz). Determining a particular LLM that potentially generated the given text is important from several perspectives: it can help to narrow down the set of LLMs for more sensitive white-box detection techniques or in cases where the generated material is harmful, misleading, or illegal, it might be useful for addressing ethical concerns and legal obligations.
 
    \paragraph{Subtask C: Changing Point Detection.} The goal of this subtask is to precisely identify the exact boundary (changing point) within a text at which the authorship transitions from a human to machine happens. The texts begin with human-written content, which at some point is automatically continued by LLMs (GPT and LlaMA series). The percentage of the human-written section varies from 0\% to 50\%. This task takes into account the fact that in many malignant use-cases of LLMs, the part of the text might be written by a human and a part might be generated by a machine. It is hard to classify a text as machine-generated if a big chunk is actually human-written. This is a way to obscure the usage of LLM, and the formulation of Subtask C addresses this challenge.
\vspace{10pt}

The task attracted a large number of participants: 126 teams for the Subtask A monolingual track, 59 teams for the Subtask A multilingual track, 70 teams for Subtask B, and 30 teams for Subtask C, with a total of 54 participating teams having submitted a system description paper for all subtasks. 

%
Next, we  introduce the MGT detection techniques considered in this shared task in \S\ref{sec:background};
    \S\ref{sec:dataset} describes the corpus and the evaluation metrics;
    \S\ref{sec:organization} details the organization of the task;
    \S\ref{sec:methods} provides an overview of the participating systems; and
    \S\ref{sec:results} discusses the evaluation results.
\section{Background}
\label{sec:background}
Detecting machine-generated text is primarily formulated as a binary classification task \cite{zellers2019defending, gehrmann-etal-2019-gltr, solaiman2019release, ippolito2019automatic}, naively distinguishing between human-written and machine-generated text. In general, there are two main approaches: the supervised methods \cite{wang2024m4gt, wang2023m4, uchendu2021turingbench, zellers2019defending, zhong2020neural, liu2022coco} and the unsupervised ones, such as zero-shot methods \cite{solaiman2019release, ippolito2019automatic, mitchell2023detectgpt, su2023detectllm, hans2024spotting}. While supervised approaches yield relatively better results, they are susceptible to overfitting \cite{mitchell2023detectgpt, su2023detectllm}. Meanwhile, unsupervised methods may require unrealistic white-box access to the generator. In the following, we provide background information on each subtask, respectively. 

\paragraph{Subtask A: Mono-lingual and Multi-lingual Binary Classification}

Given the prevalence of the binary classification task, various benchmarks assess model performance in both mono-lingual and multi-lingual settings. HC3 \cite{guo2023close} compares ChatGPT-generated text with human-written text in English and Chinese, utilizing logistic regression models trained on GLTR Test-2 features \cite{gehrmann-etal-2019-gltr} and RoBERTa \cite{liu2019roberta}-based classifiers for detection. Benchmark results by \citet{wang2023m4} include evaluations of several supervised detectors, such as RoBERTa~\cite{liu2019roberta}, XLM-R \cite{conneau2019unsupervised}, logistic regression classifier with GLTR features \cite{gehrmann2019gltr}, and stylistic features (e.g., stylometry \cite{li2014authorship}, NELA~\cite{horne2019robust} features). \citet{macko-etal-2023-multitude} create a similar resource called MULTITuDE for 11 languages in the news domain and conduct an extensive evaluation of various baselines. Our effort extends the previous works by providing evaluation setup for multiple domains, multiple languages, and for state-of-the-art LLMs, including ChatGPT and GPT-4.

\paragraph{Subtask B: Multi-Way Generator Detection}

Multi-way generator detection, attributing texts not just to their machine-generated nature but also to specific generators, resembles authorship attribution. \citet{munir2021through} find that texts from language models (LMs) have distinguishable features for source attribution. \citet{uchendu2020authorship} addresses three authorship attribution problems: (1) determining if two texts share the same origin, (2) discerning whether a text is machine or human-generated, and (3) identifying the language model responsible for text generation. Approaches like GPT-who by \citet{venkatraman2023gpt} employ UID-based features to capture unique signatures of each language model and human author, while \citet{rivera2024few} leverages representations of writing styles.

\paragraph{Subtask C: Change Point Detection}

Change point detection, which is closely tied to authorship obfuscation \cite{macko2024authorship}, extends beyond binary/multi-class classification to an adversarial co-authorship setting involving both humans and machines \cite{dugan2023real}. Machine-generated text detection methods are vulnerable to authorship obfuscation attacks such as paraphrasing \cite{crothers2022adversarial, krishna2023paraphrasing, shi2023red, koike2023outfox}, back-translation, and change point detection. Related to Subtask C, \cite{gao2024llm} introduces a dataset with mixed machine and human-written texts using operations such as polish, complete \cite{xie2023next}, rewrite \cite{shu2023rewritelm}, humanize (adding natural noise \cite{wang2021adversarial}), and adapt \cite{gero2022sparks}. \citet{kumarage2023stylometric} uses stylometric signals to quantify changes in tweets and detect when AI starts generating tweets. Different to our task, they focus on human-to-AI author changes within a given Twitter timeline.
\section{Dataset and Metrics}
\label{sec:dataset}
In this section, we describe the datasets and evaluation metrics for all subtask tracks, including the size, domains, generators, and language distribution across training, development, and test splits.

\subsection{Subtask A: Monolingual Track}
\begin{table}[t!]
    \centering
    \resizebox{\columnwidth}{!}{
            \setlength{\tabcolsep}{3pt}
    \begin{tabular}{lc|cccccc| cc}
    \toprule
        \textbf{Split} & \textbf{Source} & \textbf{\davinci} & \textbf{\chatgpt} & \textbf{Cohere} & \textbf{Dolly-v2} & \textbf{BLOOMz} & \textbf{GPT-4} & \textbf{Machine} & \textbf{Human}\\
    \midrule
      \multirow{5}{*}{Train} & Wikipedia & 3,000 & 2,995 & 2,336 & 2,702 & -- & --  & 11,033 & 14,497\\
      & Wikihow   & 3,000 & 3,000 & 3,000 & 3,000 & -- & -- &  12,000& 15,499\\
      & Reddit & 3,000 & 3,000 & 3,000 & 3,000 & -- & --& 12,000 & 15,500\\
      & arXiv & 2,999 & 3,000 & 3,000 & 3,000 &-- & --& 11,999 & 15,498\\
      & PeerRead & 2,344 & 2,344 & 2,342 & 2,344 & -- & -- & 9,374 & 2,357\\
      \midrule
    \multirow{5}{*}{Dev} & Wikipedia & -- & -- & -- & -- & 500 & -- & 500 & 500\\
      & Wikihow   & -- & -- & -- & -- & 500 & -- & 500 & 500\\
      & Reddit & -- & -- & -- & -- & 500 & -- & 500 & 500\\
      & arXiv & -- & -- & -- & -- & 500 & -- & 500 & 500\\
      & PeerRead & -- & -- & -- & -- & 500 & -- & 500 & 500\\
      \midrule
    \multirow{1}{*}{Test} & Outfox & 3,000 & 3,000 & 3,000 & 3,000 & 3,000 & 3,000 & 18,000 & 16,272\\
    \bottomrule
    \end{tabular}
    }
    \caption{\textbf{Subtasks A: Monolingual Binary Classification.} Data statistics over Train/Dev/Test splits}
    \label{tab:dataset-subtaskA-monolingual}
\end{table}

\textbf{Data:} Table~\ref{tab:dataset-subtaskA-monolingual} presents statistics across generators, domains, and splits. The training set encompasses domains such as \wikipedia, \wikihow, \reddit, \arxiv, and \peerread, comprising a total of 56,400 machine-generated and 63,351 human-written texts. \bloomz is utilized as an unseen generator in the development set, which contains 2,500 machine-generated and 2,500 human-written texts. For the test set, \outfox is introduced as the surprising domain, and \gptfour serves as the surprising generator, with a dataset of 18,000 machine-generated and 16,272 human-written texts.
\\
\textbf{Metrics:} Accuracy is used to evaluate detectors.

\subsection{Subtask A: Multilingual Track}

\begin{table}[t!]
    \centering
    \resizebox{\columnwidth}{!}{
            \setlength{\tabcolsep}{3pt}
    \begin{tabular}{lc|ccccc| cc}
    \toprule
        \textbf{Split} & \textbf{Language} & \textbf{\davinci} & \textbf{\chatgpt} & \textbf{LLaMA2} & \textbf{Jais} & \textbf{Other} & \textbf{Machine} & \textbf{Human}\\
    \midrule
      \multirow{5}{*}{Train} & English & 11,999 & 11,995 & -- & -- & 35,036 & 59,030 & 62,994\\
      & Chinese   & 2,964 & 2,970 & -- & -- & -- &  5,934 & 6,000\\
      & Urdu & -- & 2,899 & -- & -- & -- & 2,899 & 3,000\\
      & Bulgarian & 3,000 & 3,000 & -- & -- & -- &  6,000 & 6,000\\
      & Indonesian & -- & 3,000 & -- & -- & -- & 3,000 & 3,000\\
      \midrule
    \multirow{3}{*}{Dev} & Russian & 500 & 500 & -- & -- & -- &  1,000 & 1,000\\
      & Arabic  & -- & 500 & -- & -- & -- &  500 & 500\\
      & German & -- & 500 & -- & -- & -- &  500 & 500\\
      \midrule
    \multirow{4}{*}{Test} & English & 3,000 & 3,000 & -- & -- & 9,000 & 15,000 & 13,200\\
    & Arabic   & -- & 1,000 & -- & 100 & -- & 1,100 & 1,000\\
    & German & -- & 3,000 & -- & -- & -- & 3,000 & 3,000\\
    & Italian   & -- & -- & 3,000 & -- & -- & 3,000 & 3,000\\
    \bottomrule
    \end{tabular}
    }
    \caption{\textbf{Subtasks A: Multilingual Binary Classification.} Data statistics over Train/Dev/Test splits (Others generators are \cohere, \dolly and \bloomz)}
    \label{tab:dataset-subtaskA-multilingual}
\end{table}

\textbf{Data:} Table~\ref{tab:dataset-subtaskA-multilingual} presents the dataset statistics. The training set encompasses texts in English, Chinese, Urdu, Bulgarian, and Indonesian, totaling 76,863 machine-generated and 80,994 human-written texts. The development set includes Arabic (sourced from \wikipedia), Russian, and German (sourced from \wikipedia), each contributing 2,000 texts from both machine-generated and human-written sources. In the test set, Italian is introduced as the unexpected language, with \textit{\outfox} and \textit{News} serving as new domains for English, Arabic, and German texts. This set comprises 22,100 machine-generated and 20,200 human-written texts.
\\
\textbf{Metrics:} Accuracy is used to evaluate detectors.

\subsection{Subtask B}
\begin{table}[t!]
    \centering
    \resizebox{\columnwidth}{!}{
            \setlength{\tabcolsep}{3pt}
    \begin{tabular}{lc|cccccc}
    \toprule
        \textbf{Split} & \textbf{Source} & \textbf{\davinci} & \textbf{\chatgpt} & \textbf{Cohere} & \textbf{Dolly-v2} & \textbf{BLOOMz} & \textbf{Human}\\
    \midrule
      \multirow{4}{*}{Train} & Wikipedia & 3,000 & 2,995 & 2,336 & 2,702 & 2,999 & 3,000\\
      & Wikihow   & 3,000 & 3,000 & 3,000 & 3,000 & 3,000 & 2,995\\
      & Reddit & 3,000 & 3,000 & 3,000 & 3,000 & 2,999 & 3,000\\
      & arXiv & 2,999 & 3,000 & 3,000 & 3,000 & 3,000 & 2,998\\
      \midrule
    \multirow{1}{*}{Dev}
      & PeerRead & 500 & 500 & 500 & 500 & 500 & 500\\
      \midrule
    \multirow{1}{*}{Test} & Outfox & 3,000 & 3,000 & 3,000 & 3,000 & 3,000 & 3,000\\
    \bottomrule
    \end{tabular}
    }
    \caption{\textbf{Subtasks B: Multi-Way Generator Detection.} Data statistics over Train/Dev/Test splits}
    \label{tab:dataset-subtaskB}
\end{table}

\textbf{Data:} In Table~\ref{tab:dataset-subtaskB}, we incorporate texts from five generators (\davinci, \chatgpt, \cohere, \dolly, and \bloomz) alongside human-written texts. The development set features texts from the \peerread domain, while the test set introduces \outfox (specifically, student essays) as the unexpected domain.
\\
\textbf{Metrics:} Accuracy is used to evaluate detectors.

\subsection{Subtask C}
\begin{table}[t!]
    \centering
    \resizebox{\columnwidth}{!}{
    \begin{tabular}{ll | llll}
    \toprule
    Domain & Generator & Train & Dev & Test & Total \\
    \midrule
    \multirow{5}{*}{\peerread}
    & ChatGPT & \textbf{3,649 (232)} & \textbf{505 (23)} & \textbf{1,522  (89)} & 5,676 (344) \\
    & \llamatwo-7B* & 3,649 (5) & 505 (0) & \textbf{1,035 (1)} & 5,189 (6) \\
    & \llamatwo-7B & 3,649 (227) & 505 (24) & \textbf{1,522  (67)} & 5,676 (318)\\
    & \llamatwo-13B & 3,649 (192) & 505 (24) & \textbf{1,522  (84)} & 5,676 (300) \\
    & \llamatwo-70B & 3,649 (240) & 505 (21) & \textbf{1,522  (88)} & 5,676 (349) \\
    \midrule
    \multirow{5}{*}{\outfox}
    & GPT-4 & -- & -- & \textbf{1,000 (10)} & 1,000 (10) \\
    & LLaMA2-7B & -- & -- & \textbf{1,000 (8)} & 1,000 (8) \\
    & LLaMA2-13B & -- & -- & \textbf{1,000 (5)} & 1,000 (5) \\
    & LLaMA2-70B & -- & -- & \textbf{1,000 (19)} & 1,000 (19) \\
    \midrule
    Total & all & 18,245 & 2,525 & 11,123 & 31,893 \\
    \bottomrule
    \end{tabular}
    }
    \caption{\textbf{Subtask C: Change Point Detection.} We use generators GPT and \llamatwo series over domains of academic paper review (\peerread) and student essay (\outfox). The number in ``()'' is the number of examples purely generated by LLMs, i.e., human and machine boundary index=0. \llamatwo-7B* and \llamatwo-7B used different prompts. Bold data is used in shared task training, development, and test.}
    \label{tab:dataset-C}
\end{table}

\textbf{Data:}
The training and development sets for subtask C are \peerread \chatgpt generations, with \textbf{5,349}  and \textbf{505} examples respectively (first row of \tabref{tab:dataset-C}), and the test set is the combination of the \textit{test column} of \tabref{tab:dataset-C}, totaling 11,123 examples.
\\
\textbf{Metrics:}
The Mean Absolute Error (MAE) is used to evaluate the performance of the boundary detection model. It measures the average absolute difference between the predicted position index and the actual changing point.

\section{Task Organization}
\label{sec:organization}
The shared task was run in two phases:

\paragraph{Development Phase.}
Only training and development data were provided to the participants, with no gold labels available for the development set.
Participants competed against each other to achieve the best performance on the development set. A live leaderboard on CodaLab was made available to track all submissions.
Teams could make an unlimited number of submissions, and the best score for each team, regardless of the submission time, was displayed in real time on CodaLab.

\paragraph{Test Phase.} 
The test set was released, containing two additional languages—German and Italian for Subtask A Multilingual Track, generator \gptfour for the Monolingual Track, and a new domain (student essays) for Subtask B. For Subtask C, both new domains and generators were introduced (\gptfour and \llamatwo series based on \peerread and \outfox), which were not disclosed to the participants beforehand (referred to as surprise languages, domains, and generators).

Participants were given approximately three weeks to prepare their predictions. They could submit multiple runs, but they wouldn't receive feedback on their performance. Only the latest submission from each team was considered official and used for the final team ranking.

In total, 125 teams submitted results for Subtask A Monolingual, 62 for Subtask A Multilingual, 70 for Subtask B, and 30 for Subtask C. Additionally, 54 teams submitted system description papers.

After the competition concluded, we released the gold labels for both the development and test sets. Furthermore, we kept the submission system open for the test dataset for post-shared task evaluations and to monitor the state of the art across the different subtasks.
\section{Participating Systems}
\label{sec:methods}
In this section, we first summarize common features for all teams based on the information they provided in the Google Docs. Then, we delve into the methods employed by the top 3 teams, accompanied by brief descriptions of the approaches utilized by the other top 10 teams.

The approaches of all teams are presented in Appendix~\ref{sec:method-summary}.


\subsection{Monolingual Human vs Machine}
\begin{table}[t]
\adjustbox{max width=\linewidth}{
    \begin{tabular}{rccccccccc}
 Team Name & \rotatebox{90}{Ranking} & \rotatebox{90}{small PLM} & \rotatebox{90}{LLM} & \rotatebox{90}{GPT} & \rotatebox{90}{fine-tuning} & \rotatebox{90}{zero-shot} & \rotatebox{90}{few-shot (k=?)} & \rotatebox{90}{Data augmentation} & \rotatebox{90}{External Data} \\
\midrule
Genaios & 1 &  & \checkmark &  &  &  &  &  &  \\
USTC-BUPT & 2 & \checkmark &  &  & \checkmark &  &  &  &  \\
petkaz & 12 & \checkmark &  &  & \checkmark &  &  &  &  \\
HU & 17 &  & \checkmark &  & \checkmark &  &  & \checkmark &  \\
TrustAI & 20 & \checkmark &  &  & \checkmark &  &  &  &  \\
L3i++ & 25 &  & \checkmark &  & \checkmark &  &  &  &  \\
art-nat-HHU & 26 & \checkmark &  &  & \checkmark &  &  &  &  \\
Unibuc - NLP & 28 & \checkmark &  &  & \checkmark &  &  & \checkmark &  \\
NewbieML & 30 & \checkmark &  &  &  &  &  &  &  \\
QUST & 31 &  & \checkmark &  & \checkmark &  &  & \checkmark &  \\
NootNoot & 39 & \checkmark &  &  & \checkmark &  &  &  &  \\
Mast Kalandar & 40 &  & \checkmark &  & \checkmark &  &  &  &  \\
I2C-Huelva & 41 &  & \checkmark &  & \checkmark &  &  &  &  \\
Werkzeug & 45 & \checkmark &  &  & \checkmark &  &  &  &  \\
NCL-UoR & 50 &  & \checkmark &  & \checkmark &  &  &  &  \\
Sharif-MGTD & 51 &  & \checkmark &  & \checkmark & \checkmark &  &  &  \\
Collectivized Semantics & 62 & \checkmark &  &  & \checkmark &  &  &  &  \\
SINAI & 61 &  & \checkmark &  & \checkmark &  &  &  &  \\
MasonTigers & 71 & \checkmark & \checkmark &  & \checkmark & \checkmark &  &  &  \\
DUTh & 73 &  & \checkmark &  & \checkmark &  &  &  &  \\
surbhi & 74 &  & \checkmark &  & \checkmark &  &  &  &  \\
KInIT & 77 &  & \checkmark &  & \checkmark & \checkmark &  &  &  \\
RUG-D & 100 &  & \checkmark &  & \checkmark &  &  &  & \checkmark \\
RUG-5 & 101 & \checkmark & \checkmark &  & \checkmark &  &  &  &  \\
RUG-3 & 114 &  & \checkmark & \checkmark & \checkmark &  &  &  &  \\
Mashee & 115 &  & \checkmark &  &  &  & 2 &  &  \\
RUG-1 & 117 & \checkmark &  &  &  &  &  &  &  \\
\bottomrule
\end{tabular}

    }
    \caption{\textbf{Subtask A monolingual} participants methods overview. \emph{small PLM}: Pre-trained Language Model is used, \emph{LLM}: LLM is used, \emph{GPT} indicates if any GPT models are used, \emph{fine-tuning}: applying fine-tuned models, \emph{zero-shot} and \emph{few-shot (k=?)} that zero or more examples are used as demonstrations in in-context learning based on LLMs, \emph{Data augmentation} and \emph{External Data} referes to that augmented data or other external data have been used.}
    \label{tab:STA_monolingual_systems}
\end{table}

\Cref{tab:STA_monolingual_systems} provides a high-level overview of the methodologies employed by the top-ranking systems in Subtask A monolingual. 
Most systems utilized either a Pretrained Language Model (PLM) or a Large Language Model (LLM), with the majority of participants fine-tuning their models.
Usage of GPT, external data, and few-shot methods was observed in only one team each.

\textbf{Team Genaios\textsubscript{STA\_mono:1}~\citep{GenaiosSemeval2024task8}} achieved the highest performance in this subtask by extracting token-level probabilistic features (log probability and entropy) using four \llamatwo models: \llamatwo-7B, \llamatwo-7B-chat, \llamatwo-13B, and \llamatwo-13B-chat.
These features were then fed into a Transformer Encoder trained in a supervised manner.

\textbf{Team USTC-BUPT\textsubscript{STA\_mono:2}~\citep{USTC-BUPTSemeval2024task8}} secured the second position.
Their model is built upon RoBERTa, with the addition of two classification heads: one for binary classification (human or machine) using MLP layers, and another for domain classification (e.g., news, essays, etc.).
The latter is equipped with an MLP layer and a gradient reversal layer to enhance transferability between the training and test sets.
A sum-up loss is applied, resulting in approximately 8\% improvement compared to the RoBERTa baseline.

\textbf{Team PetKaz\textsubscript{STA\_mono:12}~\citep{petkazSemeval2024task8}} utilized a fine-tuned RoBERTa augmented with diverse linguistic features.

In addition to the top three teams:
\textbf{Team HU\textsubscript{STA\_mono:17}} \citep{HUSemeval2024task8} employed a contrastive learning-based approach, fine-tuning MPNet on an augmented dataset. \textbf{Team TrustAI\textsubscript{STA\_mono:20}} ensembles several classical ML classifiers, Naive Bayes, LightGBM and SGD.
\textbf{Team L3i++\textsubscript{STA\_mono:24}}~\citep{L3i++Semeval2024task8} investigated various approaches including likelihood, fine-tuning small PLMs, and LLMs, with the latter, fine-tuned \llamatwo-7B, proving to be the most effective.
\textbf{Team art-nat-HHU\textsubscript{STA\_mono:25}}~\citep{art-nat-HHU-Semeval2024task8} utilized a RoBERTa-base model combined with syntactic, lexical, probabilistic, and stylistic features.
\textbf{Team Unibuc - NLP\textsubscript{STA\_mono:28}} \citep{UnibucSemeval2024task8} jointly trained Subtasks A and B based on RoBERTa.
Most other teams fine-tuned either RoBERTa or XLM-RoBERTa for MGT detection, enhancing the models through various techniques, ranging from a mixture of experts by \textbf{Team Werkzeug\textsubscript{STA\_mono:45}}~\citep{werkzeugSemeval2024task8} to low-rank adaptation by \textbf{Team NCL-UoR\textsubscript{STA\_mono:50}} \citep{NCLUoRSemeval2024task8}, while \textbf{Team Sharif-MGTD\textsubscript{STA\_mono:51}} \citep{sharif-MGTD4ADLSemeval2024task8} preferred careful fine-tuning of PLMs alone.

\subsection{Multilingual Human vs Machine}
\begin{table}[t]
\adjustbox{max width=\linewidth}{
    \begin{tabular}{rcccccccc}
 Team Name & \rotatebox{90}{Ranking} & \rotatebox{90}{small PLM} & \rotatebox{90}{LLM} & \rotatebox{90}{GPT} & \rotatebox{90}{Fine-tuning} & \rotatebox{90}{Zero-shot}  & \rotatebox{90}{Data augmentation} & \rotatebox{90}{External Data} \\
\midrule
USTC-BUPT & 1 &  & \checkmark &  & \checkmark &  &    &  \\
FI Group & 2 & \checkmark &  &  & \checkmark &  &    &  \\
KInIT & 3 &  & \checkmark &  & \checkmark & \checkmark &    &  \\
L3i++ & 5 &  & \checkmark &  & \checkmark &  &    &  \\
QUST & 6 &  & \checkmark &  & \checkmark &  &   \checkmark &  \\
AIpom & 9 &  & \checkmark &  & \checkmark &  &    &  \\
SINAI & 21 &  & \checkmark &  & \checkmark &  &    &  \\
Unibuc-NLP & 22 & \checkmark &  &  & \checkmark &  &    &  \\
Werkzeug & 30 & \checkmark &  &  & \checkmark &  &    &  \\
RUG-5 & 32 & \checkmark & \checkmark &  & \checkmark &  &    &  \\
DUTh & 33 &  & \checkmark &  & \checkmark &  &    &  \\
RUG-D & 39 &  & \checkmark &  & \checkmark &  &    & \checkmark \\
MasonTigers & 49 & \checkmark & \checkmark &  & \checkmark & \checkmark &    &  \\
TrustAI & 55 & \checkmark &  &  & \checkmark &  &    &  \\
\bottomrule
\end{tabular}
}
\caption{\textbf{Subtask A multilingual} participants methods. }
\label{tab:STA_multilingual_systems}
\end{table}

\Cref{tab:STA_multilingual_systems} provides an overview of the methods employed by the top-performing systems for Subtask A Multilingual. Various techniques are utilized, including zero-shot learning based on LLMs, PLM-based classifiers, and ensemble models.

\textbf{Team USTC-BUPT\textsubscript{STA\_Multi:1}}~\cite{USTC-BUPTSemeval2024task8} secured the top position. They initially detect the language of the input text. For English text, they average embeddings from \llamatwo-70B, followed by classification through a two-stage CNN. For texts in other languages, the classification problem is transformed into fine-tuning a next-token prediction task using the mT5 model, incorporating special tokens for classification. Their approach integrates both monolingual and multilingual strategies, exploiting large language models for direct embedding extraction and model fine-tuning. This enables the system to adeptly handle text classification across a diverse range of languages, especially those with fewer resources.

\textbf{Team FI Group\textsubscript{STA\_Multi:2}} \cite{FIGroupSemeval2024task8} implemented a hierarchical fusion strategy that adaptively combines representations from different layers of XLM-RoBERTa-large, moving beyond the conventional "[CLS]" token classification to sequence labeling for enhanced detection of stylistic nuances.

\textbf{Team KInIT\textsubscript{STA\_Multi:3}} \cite{KInITSemeval2024task8} combined fine-tuned LLMs with zero-shot statistical methods, employing a two-step majority voting system for predictions. Their method emphasizes language identification, per-language threshold calibration, and the integration of both fine-tuned and statistical detection methods, demonstrating the power of ensemble strategies. For the LLMs, they utilized QLoRA PEFT to fine-tune Falcon-7B and Mistral-7B.

Other teams explored various approaches, like using LoRA-finetuned LLMs as classifiers (\textbf{Team AIpom\textsubscript{STA\_Multi:9}}) \cite{AIpomSemEval2024task8}, using semantic and syntactic aspects of the texts (\textbf{RFBES\textsubscript{STA\_Multi:10}}) \cite{RFBESSemeval2024task8} or fusing perplexity with text and adding a classification head (\textbf{Team SINAI\textsubscript{STA\_Multi:21}}) \cite{SINAISemeval2024task8}. Each team's method provides insights into the complexities of multilingual text detection, ranging from the use of specific LLMs and PLMs to the use of linguistic and probabilistic metrics and ensemble techniques \cite{werkzeugSemeval2024task8, RUG-DSemeval2024task8, DUThSemeval2024task8, MasonTigersSemeval2024task8, TrustAISemeval2024task8}.

\subsection{Multi-way Detection}
\begin{table}[t]
    \adjustbox{max width=\linewidth}{
        \begin{tabular}{rccccccc}
            Team Name & \rotatebox{90}{Ranking} & \rotatebox{90}{small PLM} & \rotatebox{90}{LLM} & \rotatebox{90}{GPT} & \rotatebox{90}{fine-tuning} & \rotatebox{90}{zero-shot} & \rotatebox{90}{Data augmentation} \\
            \midrule
            AISPACE & 1 &  & \checkmark &  & \checkmark &  & \checkmark \\
            Unibuc - NLP & 2 & \checkmark &  &  & \checkmark &  & \checkmark \\
            USTC-BUPT & 3 &  & \checkmark &  &  &  &  \\
            L3i++ & 6 &  & \checkmark &  & \checkmark &  &  \\
            MLab & 7 &  &  &  & \checkmark &  &  \\
            Werkzeug & 8 & \checkmark &  &  & \checkmark &  &  \\
            TrustAI & 14 & \checkmark &  &  & \checkmark &  &  \\
            MGTD4ADL & 17 &  &  &  & \checkmark &  & \checkmark \\
            scalar & 18 & \checkmark &  &  &  &  & \checkmark \\
            UMUT & 23 & \checkmark &  &  & \checkmark &  &  \\
            QUST & 36 &  & \checkmark &  & \checkmark &  & \checkmark \\
            MasonTigers & 38 & \checkmark & \checkmark &  & \checkmark & \checkmark &  \\
            RUG-5 & 41 & \checkmark & \checkmark &  & \checkmark &  &  \\
            RUG-D & 44 &  & \checkmark &  & \checkmark &  &  \\
            DUTh & 49 &  & \checkmark &  & \checkmark &  &  \\
            clulab-UofA & 62 &  & \checkmark & \checkmark & \checkmark &  & \checkmark \\
            \bottomrule
            \end{tabular}
        }
        \caption{\textbf{Subtask B} Participants method overview.}
        \label{tab:STB_systems}
    \end{table}

\tabref{tab:STB_systems} provides an overview of the approaches employed by the top-ranking systems for Subtask B. Similar to Subtask A, most solutions do not use GPT and zero-shot approaches. The best-performing solutions primarily exploit LLMs and data augmentation.

\textbf{Team AISPACE\textsubscript{STB:1}} \cite{AISPACESemeval2024task8} achieved the highest performance in this subtask by fine-tuning various encoder and encoder-decoder models, including RoBERTa, DeBERTa, XLNet, Longformer, and T5. They augmented the data with instances from Subtask A and explored the effects of different loss functions and learning rate values. Their method leverages a weighted Cross-Entropy loss to balance samples in different classes and uses an ensemble of fine-tuned models to improve robustness.

\textbf{Team Unibuc - NLP\textsubscript{STB:2}} \cite{UnibucSemeval2024task8} employed a Transformer-based model with a unique two-layer feed-forward network as a classification head. They also augmented the data with instances from the Subtask A monolingual dataset.

\textbf{Team USTC-BUPT\textsubscript{STB:3}} \cite{USTC-BUPTSemeval2024task8} leveraged \llamatwo-70B to obtain token embeddings and applied a three-stage classification. They first distinguished human-generated from machine-generated text using \llamatwo-70B, then categorized \chatgpt and \cohere as one class and distinguished them from \davinci, \bloomz, and \dolly. Finally, they performed binary classification between ChatGPT and Cohere.

\textbf{Team L3i++\textsubscript{STB:6}} \cite{L3i++Semeval2024task8} conducted a comparative study among three groups of methods: metric-based models, fine-tuned classification language models (RoBERTa, XLM-R), and a fine-tuned LLM, \llamatwo-7B, finding \llamatwo to outperform other methods. They analyzed errors and various factors in their paper.

\textbf{Team MLab\textsubscript{STB:7}} \cite{MLabSemeval2024task8} fine-tuned DeBERTa and analyzed the embeddings from the last layer to provide insights into the embedding space of the model.

\textbf{Team Werkzeug\textsubscript{STB:8}} \cite{werkzeugSemeval2024task8} utilized RoBERTa-large and XLM-RoBERTa-large to encode text, addressing the problem of anisotropy in text embeddings produced by pre-trained language models (PLMs) by introducing a learnable parametric whitening (PW) transformation. They also used multiple PW transformation layers as experts under the mixture-of-experts (MoE) architecture to capture features of LLM-generated text from different perspectives.

Other teams explored various approaches, including different loss functions and sentence transformers (\textbf{Team MGTD4ADL\textsubscript{STB:17}}) \cite{MGTD4ADLSemeval2024task8}, RoBERTa fine-tuning (\textbf{Team UMUTeam\textsubscript{STB:23}}) \cite{UMUTSemeval2024task8}, stacking ensemble techniques (\textbf{Team MasonTigers\textsubscript{STB:38}}) \cite{MasonTigersSemeval2024task8}, and basic ML models with linguistic-stylistic features (\textbf{Team RUG-5\textsubscript{STB:41}}) \cite{RUG-5Semeval2024task8}.

\subsection{Boundary Identification}
\begin{table}[t]
\adjustbox{max width=\linewidth}{
\begin{tabular}{rccccccc}
Team Name & \rotatebox{90}{Ranking} & \rotatebox{90}{small PLM} & \rotatebox{90}{LLM} & \rotatebox{90}{LSTM (+) CNN} & \rotatebox{90}{fine-tuning} & \rotatebox{90}{Data augmentation} & \rotatebox{90}{CRF layer} \\
\midrule
TM-TREK\textsubscript{STC:1} & 1 & \checkmark & &  & \checkmark & & \checkmark \\
AIpom\textsubscript{STC:2} & 2 & & \checkmark & & \checkmark & & \\
USTC-BUPT\textsubscript{STC:3} & 3 & \checkmark & \checkmark & & \checkmark & \checkmark & \\
RKadiyala\textsubscript{STC:6} & 6 & \checkmark & & & \checkmark & & \checkmark \\
DeepPavlov\textsubscript{STC:7} & 7 & \checkmark & & & \checkmark & \checkmark & \\
RUG-5 \textsubscript{STC:17} & 17 & \checkmark & \checkmark & & \checkmark & & \\
TueCICL\textsubscript{STC:22} & 22 & & & \checkmark & & & \\
jelarson\textsubscript{STC:25} & 25 & & & & & & \\
MasonTigers \textsubscript{STC:27} & 27 & \checkmark & & & & & \\
Unibuc-NLP\textsubscript{STC:28} & 28 & & & \checkmark & \checkmark & & \\
\bottomrule
\end{tabular}
}
\caption{\textbf{Subtask C} Participants method overview.}
\label{tab:STC_systems}
\end{table}

\tabref{tab:STC_systems} presents an overview of the methods used by the top-ranking systems for Subtask C. The best performing solutions are mainly based on ensemble strategies, with some employing data augmentation.

\textbf{Team TM-TREK\textsubscript{STC:1}} \cite{TM-TREKSemEval2024task8} achieved the highest performance in Subtask C. They utilized an ensemble of XLNet models, each trained with a distinct seed, and used a straightforward voting mechanism on the output logits. They also explored the integration of LSTM and CRF layers on top of various PLMs, along with continued pretraining and fine-tuning of PLMs, and dice loss functions to enhance model performance.

\textbf{Team AIpom\textsubscript{STC:2}} \cite{AIpomSemEval2024task8} introduced a novel two-stage pipeline merging outputs from an instruction-tuned, decoder-only (Mistral-7B-OpenOrca) model and two encoder-only sequence taggers.

\textbf{Team USTC-BUPT\textsubscript{STC:3}} \cite{USTC-BUPTSemeval2024task8} fine-tuned a DeBERTa model with data augmentation and framed the task as a token classification problem.

\textbf{Team RKadiyala \textsubscript{STC:6}} \cite{RkadiyalaSemeval2024task8} fine-tuned various encoder-based models with a Conditional Random Field (CRF) layer and found Deberta-V3 to perform the best on the development set.

\textbf{Team DeepPavlov \textsubscript{STC:7}} \cite{DeepPavlovSemeval2024task8} fine-tuned the Deberta-v3 model using the provided dataset and developed a data preprocessing pipeline for data augmentation.

Other teams explored diverse CNN, LSTM (\textbf{Team TueCICL \textsubscript{STC:22}}) \cite{tueCICLSemeval2024task8}, (\textbf{Team Unibuc - NLP \textsubscript{STC:28}}) \cite{UnibucSemeval2024task8}, and regression-based (\textbf{Team jelarson \textsubscript{STC:25}}) \cite{jelarson678Semeval2024task8} techniques to address this challenge, although many did not surpass the baselines due to issues related to model overfitting or inadequate word embeddings.

\section{Results and Discussion}
\label{sec:results}
\begin{table*}[t!]
    \centering
    \resizebox{\textwidth}{!}{
    \begin{tabular}{cl|ccc|c}
        \toprule
        Rank & Team & Prec & Recall & F1-score & Acc \\
        \midrule
* & dianchi & 96.21 & 99.19 & 97.68 & 97.53 \\
1 & Genaios & 96.11 & 98.03 & 97.06 & 96.88 \\
2 & USTC-BUPT & 95.75 & 96.86 & 96.30 & 96.10 \\
3 & mail6djj & 94.87 & 97.18 & 96.02 & 95.76 \\
4 & howudoin & 93.48 & 98.12 & 95.74 & 95.42 \\
5 & idontknow & 94.57 & 95.42 & 94.99 & 94.72 \\
6 & seven & 90.12 & 98.31 & 94.04 & 93.46 \\
7 & zongxiong & 93.54 & 93.82 & 93.68 & 93.35 \\
8 & mahsaamani & 90.59 & 96.23 & 93.32 & 92.77 \\
9 & bennben & 91.49 & 95.05 & 93.24 & 92.76 \\
10 & infinity2357 & 91.92 & 90.96 & 91.43 & 91.05 \\
11 & AISPACE & 84.76 & 99.92 & 91.72 & 90.52 \\
12 & petkaz & 85.54 & 98.59 & 91.61 & 90.51 \\
13 & moniszcz1 & 86.96 & 95.68 & 91.11 & 90.20 \\
14 & moniszcz3 & 86.96 & 95.68 & 91.11 & 90.20 \\
15 & flash & 82.39 & 99.77 & 90.25 & 88.68 \\
* & baseline & 93.36 & 84.02 & 88.44 & 88.47 \\
16 & ericmxf & 81.71 & 99.98 & 89.93 & 88.24 \\
17 & HU & 82.63 & 97.24 & 89.34 & 87.81 \\
18 & jrutkowski2 & 84.58 & 93.44 & 88.79 & 87.61 \\
19 & lihaoran & 89.26 & 85.86 & 87.53 & 87.15 \\
20 & TrustAI & 89.21 & 85.50 & 87.31 & 86.95 \\
21 & TM-TREK & 79.47 & 99.99 & 88.56 & 86.43 \\
22 & jojoc & 86.30 & 87.76 & 87.02 & 86.25 \\
23 & RFBES & 91.58 & 80.64 & 85.76 & 85.95 \\
24 & L3i++ & 81.41 & 94.66 & 87.53 & 85.84 \\
25 & art-nat-HHU & 86.29 & 86.04 & 86.17 & 85.49 \\
26 & FI Group & 79.52 & 96.99 & 87.39 & 85.30 \\
27 & phuhoang & 87.72 & 83.65 & 85.64 & 85.26 \\
28 & Unibuc-NLP & 78.01 & 99.86 & 87.59 & 85.14 \\
29 & sushvin & 82.65 & 89.76 & 86.06 & 84.73 \\
30 & NewbieML & 79.32 & 95.06 & 86.48 & 84.39 \\
31 & QUST & 76.88 & 99.91 & 86.89 & 84.17 \\
32 & MLab & 83.17 & 85.83 & 84.48 & 83.44 \\
33 & ziweizheng & 82.31 & 85.01 & 83.63 & 82.53 \\
34 & AIpom & 74.34 & 99.97 & 85.27 & 81.86 \\
35 & lyaleo & 79.38 & 87.82 & 83.39 & 81.62 \\
36 & yunhfang & 75.26 & 95.31 & 84.10 & 81.08 \\
37 & sankalpbahad & 78.22 & 88.08 & 82.86 & 80.86 \\
38 & aktsvigun & 78.22 & 88.08 & 82.86 & 80.86 \\
39 & NootNoot & 78.22 & 88.08 & 82.86 & 80.86 \\
40 & Mast Kalandar  & 74.65 & 96.16 & 84.05 & 80.83 \\
41 & I2C-Huelva & 73.92 & 98.01 & 84.28 & 80.79 \\
42 & priority497 & 73.31 & 99.69 & 84.49 & 80.78 \\
43 & wjm123 & 73.31 & 99.69 & 84.49 & 80.78 \\
44 & scalar & 73.10 & 99.97 & 84.45 & 80.67 \\
45 & werkzeug & 75.28 & 93.88 & 83.56 & 80.59 \\
46 & blain & 72.51 & 99.96 & 84.05 & 80.07 \\
47 & xxm981215 & 72.32 & 99.79 & 83.86 & 79.83 \\
48 & moyanxinxu & 72.32 & 99.79 & 83.86 & 79.83 \\
49 & jrutkowskikag1 & 73.02 & 97.54 & 83.52 & 79.78 \\
50 & NCL-UoR & 75.10 & 90.84 & 82.22 & 79.37 \\
51 & Sharif-MGTD & 73.41 & 93.75 & 82.35 & 78.89 \\
52 & wgm123 & 71.16 & 99.94 & 83.13 & 78.69 \\
53 & logiczmaksimka & 70.79 & 99.29 & 82.65 & 78.11 \\
54 & somerandomjj & 70.43 & 98.55 & 82.15 & 77.51 \\
55 & totylkokuba & 70.43 & 98.55 & 82.15 & 77.51 \\
56 & lly123 & 69.25 & 99.95 & 81.81 & 76.66 \\
57 & mimkag2 & 69.69 & 97.99 & 81.45 & 76.56 \\
58 & priyansk & 69.28 & 99.36 & 81.64 & 76.53 \\
59 & nampfiev1995 & 77.92 & 76.93 & 77.43 & 76.44 \\
60 & xiangrunli & 68.23 & 99.97 & 81.11 & 75.54 \\
61 & roywang & 68.23 & 99.97 & 81.11 & 75.54 \\
* & badrock & 71.13 & 89.50 & 79.27 & 75.41 \\
\bottomrule
\end{tabular}


    \begin{tabular}{|cl|ccc|c}
        \toprule
        Rank & Team & Prec & Recall & F1-score & Acc \\
        \midrule
62 & Collectivized Semantics & 68.21 & 99.39 & 80.90 & 75.35 \\
63 & IUCL & 68.13 & 98.33 & 80.49 & 74.96 \\
64 & annedadaa & 68.01 & 97.69 & 80.19 & 74.66 \\
65 & cmy99 & 67.92 & 97.96 & 80.22 & 74.62 \\
66 & xiaoll & 67.92 & 97.96 & 80.22 & 74.62 \\
67 & SINAI & 67.31 & 99.88 & 80.42 & 74.46 \\
68 & yuwert777 & 68.78 & 92.96 & 79.06 & 74.14 \\
69 & yaoxy & 68.78 & 92.96 & 79.06 & 74.14 \\
70 & moniszcz & 67.25 & 97.66 & 79.65 & 73.79 \\
71 & MasonTigers & 67.59 & 95.72 & 79.23 & 73.64 \\
72 & AT & 67.30 & 96.59 & 79.33 & 73.56 \\
73 & DUTh & 66.27 & 99.92 & 79.69 & 73.24 \\
74 & surbhi & 69.38 & 87.40 & 77.35 & 73.12 \\
75 & thanet & 69.47 & 86.69 & 77.13 & 73.00 \\
76 & Kathlalu & 74.47 & 73.89 & 74.18 & 72.98 \\
77 & KInIT & 66.14 & 98.44 & 79.12 & 72.71 \\
78 & iimasNLP & 67.81 & 87.08 & 76.25 & 71.50 \\
79 & wwzzhh & 64.38 & 99.49 & 78.18 & 70.82 \\
80 & bharathsk & 64.48 & 98.69 & 78.00 & 70.76 \\
81 & apillay2 & 64.48 & 98.69 & 78.00 & 70.76 \\
82 & ashinee20 & 71.91 & 71.57 & 71.74 & 70.39 \\
83 & longfarmer & 63.81 & 99.03 & 77.61 & 69.99 \\
84 & mlnick & 63.61 & 100 & 77.76 & 69.96 \\
85 & vasko & 63.47 & 99.93 & 77.63 & 69.75 \\
86 & Groningen F & 72.74 & 65.62 & 68.99 & 69.02 \\
87 & hhy123 & 62.76 & 99.94 & 77.11 & 68.83 \\
88 & 1024m & 62.54 & 99.98 & 76.95 & 68.53 \\
89 & lhy123 & 62.54 & 99.96 & 76.94 & 68.53 \\
90 & thang & 62.25 & 99.98 & 76.73 & 68.14 \\
91 & nikich28 & 61.90 & 99.26 & 76.25 & 67.52 \\
92 & niceone & 61.70 & 98.39 & 75.84 & 67.08 \\
93 & pmalesa & 60.73 & 97.87 & 74.95 & 65.64 \\
94 & mahaalblooki & 60.85 & 96.23 & 74.56 & 65.51 \\
95 & bertsquad & 60.32 & 98.94 & 74.95 & 65.27 \\
96 & jjonczyk & 60.25 & 99.31 & 75.00 & 65.23 \\
97 & dkoterwa & 60.12 & 99.98 & 75.09 & 65.16 \\
98 & lystsoval & 92.30 & 35.31 & 51.07 & 64.48 \\
99 & sunilgundapu & 59.22 & 99.31 & 74.20 & 63.72 \\
100 & RUG-D & 59.19 & 99.35 & 74.18 & 63.68 \\
101 & RUG-5 & 60.79 & 84.13 & 70.58 & 63.17 \\
102 & harshul24 & 54.55 & 57.14 & 55.81 & 62.00 \\
103 & basavraj10 & 54.55 & 57.14 & 55.81 & 62.00 \\
104 & samnlptaskab & 58.01 & 99.93 & 73.41 & 61.97 \\
105 & partnlu & 57.87 & 99.96 & 73.31 & 61.76 \\
106 & teams2024 & 57.78 & 99.97 & 73.23 & 61.61 \\
107 & Rkadiyala & 57.30 & 99.98 & 72.85 & 60.86 \\
108 & rtuora & 57.18 & 99.89 & 72.73 & 60.66 \\
109 & teamlanlp2 & 56.93 & 99.71 & 72.48 & 60.24 \\
110 & jakubbebacz & 56.89 & 99.88 & 72.49 & 60.18 \\
111 & dandread & 56.19 & 99.87 & 71.92 & 59.04 \\
112 & pask1 & 55.97 & 99.80 & 71.72 & 58.67 \\
113 & skillissue & 55.14 & 100 & 71.09 & 57.27 \\
114 & RUG-3 & 54.92 & 99.73 & 70.83 & 56.87 \\
115 & Mashee & 57.11 & 59.58 & 58.32 & 55.27 \\
116 & TueCICL & 55.37 & 69.61 & 61.68 & 54.57 \\
117 & RUG-1 & 52.52 & 100 & 68.87 & 52.52 \\
118 & novice8 & 52.24 & 68.68 & 59.34 & 50.57 \\
119 & ronghaopan & 52.49 & 38.47 & 44.40 & 48.70 \\
120 & kamer & 52.89 & 29.56 & 37.93 & 48.48 \\
121 & helenpy & 75.29 & 1.79 & 3.51 & 48.11 \\
122 & ascisel & 7.14 & 0.01 & 0.01 & 47.44 \\
123 & laida & 40.18 & 19.28 & 26.06 & 42.53 \\
124 & nz28555 & 40.31 & 33.18 & 36.40 & 39.10 \\
&&&&&\\

\bottomrule
\end{tabular}

}
    \caption{\textbf{Subtask A monolingual} Prec (precision), Recall, and F1-scores(\%) with respect to \textbf{MGT}.
    }
    \label{tab:subtaskA_monolingual}
\end{table*}




\subsection{Subtask A}
There were three submissions for subtask A, which were submitted in time, but had the wrong file name, which prevented us from scoring them automatically. We eventually manually fixed the names and scored them, and we also added them to the ranking but marked them with a *. They should be considered as unofficial submissions.

\paragraph{Monolingual}

\begin{table}[t!]
    \centering
    \resizebox{0.95\columnwidth}{!}{
    \begin{tabular}{cl|ccc|c}
        \toprule
        Rank & Team & Prec & Recall & F1-score & Acc \\
        \midrule
1 & USTC-BUPT & 94.93 & 97.53 & 96.21 & 95.99 \\
2 & FI Group & 94.28 & 98.00 & 96.10 & 95.85 \\
3 & KInIT & 92.95 & 97.86 & 95.34 & 95.00 \\
4 & priyansk & 90.70 & 98.14 & 94.28 & 93.77 \\
5 & L3i++ & 92.47 & 94.00 & 93.23 & 92.87 \\
6 & QUST & 90.45 & 90.98 & 90.71 & 90.27 \\
7 & xxm981215 & 90.45 & 90.98 & 90.71 & 90.27 \\
8 & NCL-UoR & 81.42 & 95.41 & 87.86 & 86.23 \\
9 & AIpom & 80.72 & 95.80 & 87.61 & 85.85 \\
10 & RFBES & 85.43 & 85.27 & 85.35 & 84.71 \\
11 & blain & 76.12 & 98.67 & 85.94 & 83.14 \\
12 & xiangrunli & 75.20 & 99.67 & 85.73 & 82.66 \\
13 & wgm123 & 75.20 & 99.67 & 85.73 & 82.66 \\
14 & roywang & 75.08 & 99.75 & 85.68 & 82.58 \\
15 & logiczmaksimka & 74.34 & 99.33 & 85.04 & 81.74 \\
16 & zaratiana & 74.75 & 96.68 & 84.31 & 81.21 \\
17 & thanet & 76.18 & 92.56 & 83.58 & 81.00 \\
* & baseline & 73.45 & 99.30 & 84.44 & 80.89 \\
18 & cmy99 & 73.29 & 99.61 & 84.45 & 80.83 \\
19 & lly123 & 73.09 & 99.67 & 84.33 & 80.65 \\
20 & moyanxinxu & 73.09 & 99.67 & 84.33 & 80.65 \\
21 & SINAI & 72.51 & 99.91 & 84.04 & 80.17 \\
22 & Unibuc-NLP & 71.82 & 99.79 & 83.52 & 79.43 \\
23 & annedadaa & 72.16 & 98.57 & 83.32 & 79.39 \\
24 & 1024m & 71.03 & 99.91 & 83.03 & 78.66 \\
25 & sunilgundapu & 71.04 & 98.86 & 82.67 & 78.35 \\
26 & hirak & 70.79 & 99.66 & 82.78 & 78.34 \\
27 & bertsquad & 70.45 & 99.12 & 82.36 & 77.82 \\
28 & Rkadiyala & 69.99 & 99.94 & 82.33 & 77.59 \\
* & dianchi & 69.88 & 99.91 & 82.24 & 77.46 \\
29 & lyaleo & 69.50 & 99.83 & 81.95 & 77.03 \\
30 & werkzeug & 69.33 & 99.81 & 81.82 & 76.83 \\
31 & mlnick & 69.25 & 99.81 & 81.77 & 76.75 \\
32 & RUG-5 & 69.90 & 96.78 & 81.17 & 76.55 \\
33 & DUTh & 68.95 & 99.93 & 81.60 & 76.45 \\
34 & dandread & 68.31 & 99.75 & 81.09 & 75.69 \\
35 & Genaios & 68.30 & 99.73 & 81.07 & 75.67 \\
36 & vasko & 67.99 & 98.68 & 80.50 & 75.03 \\
37 & thang & 67.16 & 99.78 & 80.29 & 74.40 \\
38 & mahsaamani & 68.53 & 93.21 & 78.98 & 74.09 \\
39 & RUG-D & 65.03 & 99.55 & 78.67 & 71.79 \\
40 & omarnasr & 64.80 & 99.21 & 78.40 & 71.43 \\
41 & lhy123 & 64.62 & 99.85 & 78.46 & 71.36 \\
42 & priority497 & 64.62 & 99.85 & 78.46 & 71.36 \\
43 & hhy123 & 64.47 & 99.84 & 78.35 & 71.17 \\
44 & wjm123 & 64.47 & 99.84 & 78.35 & 71.17 \\
45 & aktsvigun & 62.83 & 99.36 & 76.98 & 68.96 \\
46 & sankalpbahad & 62.83 & 99.36 & 76.98 & 68.96 \\
47 & NootNoot & 62.83 & 99.36 & 76.98 & 68.96 \\
48 & nampfiev1995 & 61.37 & 77.15 & 68.36 & 62.70 \\
49 & MasonTigers & 56.77 & 100 & 72.42 & 60.21 \\
50 & RUG-1 & 51.33 & 97.39 & 67.23 & 51.22 \\
51 & novice8 & 51.95 & 84.56 & 64.36 & 51.08 \\
52 & scalar & 52.04 & 80.17 & 63.11 & 51.04 \\
53 & mahaalblooki & 48.96 & 51.24 & 50.08 & 50.55 \\
54 & Sharif-MGTD & 51.42 & 67.13 & 58.23 & 50.53 \\
55 & TrustAI & 51.15 & 62.04 & 56.07 & 50.06 \\
56 & sky2024just & 51.96 & 26.37 & 34.99 & 48.79 \\
57 & laida & 48.88 & 20.56 & 28.94 & 47.27 \\
\bottomrule
\end{tabular}

}
    \caption{\textbf{Subtask A multilingual} Prec (precision), Recall, and F1-scores(\%) with respect to \textbf{MGT}.}
    \label{tab:subtaskA_multilingual}
\end{table}

\begin{table}[t!]
    \centering
    \resizebox{0.93\columnwidth}{!}{
    \begin{tabular}{cl|ccc|c}
        \toprule
        Rank & Team & Prec & Recall & F1-score & Acc \\
        \midrule
1 & AISPACE & 91.81 & 90.85 & 90.84 & 90.85 \\
2 & Unibuc - NLP & 88.69 & 86.96 & 87.03 & 86.96 \\
3 & USTC-BUPT & 89.54 & 84.33 & 82.72 & 84.33 \\
4 & dianchi & 86.45 & 83.48 & 83.62 & 83.48 \\
5 & NootNoot & 86.68 & 83.12 & 83.15 & 83.12 \\
6 & L3i++ & 86.01 & 83.12 & 83.08 & 83.12 \\
7 & MLab & 85.00 & 82.67 & 82.76 & 82.67 \\
8 & werkzeug & 86.30 & 82.23 & 81.63 & 82.23 \\
9 & flash & 88.29 & 82.23 & 79.46 & 82.23 \\
10 & juse7198 & 86.83 & 82.03 & 80.72 & 82.03 \\
11 & idontknow & 88.44 & 80.94 & 77.47 & 80.94 \\
12 & TM-TREK & 86.42 & 79.84 & 79.46 & 79.84 \\
13 & howudoin & 80.02 & 79.68 & 79.79 & 79.68 \\
14 & TrustAI & 83.80 & 79.19 & 79.07 & 79.19 \\
15 & I2C-Huelva & 84.45 & 78.90 & 78.82 & 78.90 \\
16 & ericmxf & 85.52 & 78.74 & 76.88 & 78.74 \\
17 & MGTD4ADL & 83.78 & 76.96 & 74.46 & 76.96 \\
18 & scalar & 81.90 & 76.26 & 76.00 & 76.26 \\
19 & ronghaopan & 81.11 & 75.19 & 71.38 & 75.35 \\
20 & sunilgundapu & 81.06 & 75.06 & 73.81 & 75.06 \\
* & baseline & 81.14 & 74.61 & 72.59 & 74.61 \\
21 & Collectivized Semantics & 82.35 & 73.87 & 70.26 & 73.87 \\
22 & priyansk & 78.06 & 73.36 & 67.05 & 73.36 \\
23 & logiczmaksimka & 67.73 & 69.13 & 64.36 & 69.13 \\
24 & annedadaa & 79.55 & 68.98 & 64.55 & 68.98 \\
25 & hhy123 & 65.94 & 67.77 & 63.01 & 67.77 \\
26 & xiangrunli & 65.94 & 67.77 & 63.01 & 67.77 \\
27 & wjm123 & 65.94 & 67.77 & 63.01 & 67.77 \\
28 & lhy123 & 65.94 & 67.77 & 63.01 & 67.77 \\
29 & lly123 & 65.94 & 67.77 & 63.01 & 67.77 \\
30 & wgm123 & 65.94 & 67.77 & 63.01 & 67.77 \\
31 & moyanxinxu & 65.94 & 67.77 & 63.01 & 67.77 \\
32 & priority497 & 65.94 & 67.77 & 63.01 & 67.77 \\
33 & thang & 66.36 & 67.68 & 63.79 & 67.68 \\
34 & blain & 63.15 & 67.23 & 62.35 & 67.23 \\
35 & xxm981215 & 65.77 & 67.21 & 62.41 & 67.21 \\
36 & QUST & 65.77 & 67.21 & 62.41 & 67.21 \\
37 & mahaalblooki & 63.72 & 66.27 & 61.82 & 66.27 \\
38 & MasonTigers & 73.62 & 65.04 & 64.47 & 65.04 \\
39 & Rkadiyala & 65.81 & 64.91 & 59.98 & 64.91 \\
40 & 1024m & 66.10 & 64.38 & 59.82 & 64.38 \\
41 & RUG-5 & 62.21 & 64.21 & 59.04 & 64.21 \\
42 & thanet & 63.42 & 61.88 & 55.58 & 61.88 \\
43 & mlnick & 66.84 & 61.79 & 57.53 & 61.79 \\
44 & RUG-D & 66.39 & 61.54 & 53.82 & 61.54 \\
45 & Groningen F & 60.10 & 60.84 & 57.90 & 60.84 \\
46 & NCL-UoR & 69.03 & 60.15 & 58.05 & 60.15 \\
47 & mahsaamani & 60.41 & 59.42 & 52.89 & 59.42 \\
48 & dandread & 71.95 & 58.35 & 52.28 & 58.35 \\
49 & DUTh & 63.71 & 56.68 & 51.25 & 56.68 \\
50 & bertsquad & 57.27 & 55.97 & 51.49 & 55.97 \\
51 & RUG-3 & 61.51 & 54.23 & 49.26 & 54.23 \\
52 & cmy99 & 58.04 & 53.35 & 50.58 & 53.35 \\
53 & skysky12 & 60.86 & 53.31 & 50.14 & 53.31 \\
54 & vasko & 59.98 & 52.82 & 50.38 & 52.82 \\
55 & phuhoang & 61.54 & 50.79 & 50.21 & 50.79 \\
56 & rtuora & 54.21 & 50.32 & 44.15 & 50.32 \\
57 & AT & 53.13 & 48.59 & 43.91 & 48.59 \\
58 & teams2024 & 45.50 & 47.01 & 41.05 & 47.01 \\
59 & windwind22 & 39.87 & 39.31 & 32.79 & 39.31 \\
60 & helenpy & 39.88 & 38.27 & 32.20 & 38.27 \\
61 & iimasNLP & 39.88 & 38.27 & 32.20 & 38.27 \\
62 & clulab-UofA & 37.53 & 29.29 & 24.58 & 29.29 \\
63 & samnlptaskab & 25.78 & 27.81 & 21.07 & 27.81 \\
64 & mhr2004 & 17.06 & 17.06 & 17.06 & 17.06 \\
65 & xiaoll & 5.73 & 17.15 & 8.47 & 17.02 \\
66 & surbhi & 17.24 & 16.77 & 15.10 & 16.77 \\
67 & roywang & 2.78 & 16.67 & 4.76 & 16.67 \\
68 & RUG-1 & 2.78 & 16.67 & 4.76 & 16.67 \\
69 & novice8 & 16.39 & 16.55 & 13.93 & 16.55 \\
70 & NewbieML & 15.99 & 15.58 & 14.13 & 15.30 \\
\bottomrule
\end{tabular}

}
    \caption{\textbf{Subtask B: Multi-Way Generator Detection} Prec (precision), Recall, and F1-scores(\%) macro average.
    }
    \label{tab:subtaskB-results}
\end{table}

\begin{table}[t!]
    \centering
    \resizebox{\columnwidth}{!}{
    \begin{tabular}{cl|c}
        \toprule
        Rank & Team & MAE \\
        \midrule
1 & TM-TREK & 15.68 \\
2 & AIpom & 15.94 \\
3 & USTC-BUPT & 17.70 \\
4 & ywnh111 & 18.08 \\
5 & ywnh222 & 18.51 \\
6 & Rkadiyala & 18.54 \\
7 & DeepPavlov & 19.25 \\
8 & knk42 & 19.42 \\
9 & vasko & 19.93 \\
10 & logiczmaksimka & 19.93 \\
11 & AISPACE & 21.19 \\
* & baseline & 21.54 \\
12 & ericmxf & 21.55 \\
13 & blain & 21.80 \\
14 & 1024m & 22.36 \\
15 & cmy99 & 24.68 \\
\bottomrule
\end{tabular}
\begin{tabular}{|cl|c}
        \toprule
        Rank & Team & MAE \\
        \midrule

16 & mahaalblooki & 25.95 \\
17 & RUG-5 & 26.07 \\
18 & mahsaamani & 26.27 \\
19 & aktsvigun & 26.40 \\
20 & skillissue & 27.99 \\
21 & NootNoot & 28.01 \\
22 & TueCICL & 34.88 \\
23 & dandread & 35.17 \\
24 & novice8 & 44.82 \\
25 & jelarson & 48.14 \\
26 & TueSents & 58.95 \\
27 & MasonTigers & 60.78 \\
28 & Unibuc - NLP & 74.28 \\
29 & lanileqiu & 78.18 \\
30 & scalar & 87.72 \\
&&\\
\bottomrule
\end{tabular}

}
    \caption{\textbf{Subtask C: Boundary Identification}.
    }
    \label{tab:subtaskC-results}
\end{table}

\tabref{tab:subtaskA_monolingual} presents the performance of systems in the monolingual track of Subtask A. Out of 125 participating teams, 15 surpassed the baseline, with the top-performing team (Genaios) achieving an accuracy of 96.88. Notably, several teams demonstrated high precision and recall scores, indicating robust performance in distinguishing between human-generated and machine-generated text in a binary classification context.

\paragraph{Multilingual}

\autoref{tab:subtaskA_multilingual} presents the performance of systems in the multilingual track of Subtask A, where Team USTC-BUPT emerges as the top performer among 62 participating teams, achieving an accuracy of 95.99, remarkably close to the English-only result. Their methodology entails a blend of language detection and fine-tuning tasks using \llamatwo-70B for English and the mT5 model for others, showcasing their adaptability across diverse languages.

Similarly, among the 22 teams surpassing the baseline, the majority leverage advanced LLMs such as LLaMA, Mistral, etc., while also emphasizing syntax and writing style differences between human and AI-generated texts. For example, Team FI Group implements a hierarchical fusion strategy to adaptively fuse representations from different BERT layers, prioritizing syntax over semantics for improved classification accuracy. Likewise, Team KInIT employs an ensemble approach, combining fine-tuned LLMs with zero-shot statistical methods, resulting in a unique combination of techniques that effectively enhances classification accuracy.

Overall, these successful methodologies underscore the importance of leveraging advanced LLMs, ensemble techniques, and comprehensive analysis to achieve superior performance in detecting machine-generated text across multilingual contexts. 


\subsection{Subtask B}

For Subtask B (Multi-Way detection), 70 teams participated, with 20 surpassing the baseline of 74.61 accuracy. \tabref{tab:subtaskB-results} displays the full results. In summary, the subtask results underline the effectiveness of diverse and innovative approaches, including fine-tuning advanced models (e.g., RoBERTa, DeBERTa, XLNet, Longformer, T5), data augmentation (e.g., using Subtask A instances), ensemble strategies, and the exploration of novel loss functions and learning techniques. The leading entries showcased a range of methodologies, from leveraging the power of large language models and addressing embedding anisotropy to integrating traditional and neural methods, underscoring the dynamic and evolving nature of NLP research. For instance, Team AISPACE utilized a weighted Cross-Entropy loss and an ensemble approach based on model performance per class, which led to the highest accuracy of 90.85.

\subsection{Subtask C}

Of the 30 systems that were submitted for Subtask C, 11 outperformed the baseline MAE of 21.54. The top system, TM-TREK, achieved the best submitted MAE of 15.68.
A significant majority of the top-performing teams relied on ensembles of PLMs, indicating a consensus that combining the strengths of multiple models can lead to more robust and accurate predictions. This approach leverages the diverse representations and strengths of different models to mitigate weaknesses inherent in individual systems.

Data augmentation emerged as a critical strategy among leading teams, suggesting its effectiveness in enhancing model performance by providing a richer, more varied training dataset. This includes both the generation of new training examples and the manipulation of existing data to better capture the complexity and variability of natural language.

Despite the advanced methodologies deployed, some teams struggled with issues related to overfitting and the adequacy of word embeddings. This underscores the ongoing challenges in developing models that generalize well to unseen data and the critical role of embeddings in capturing semantic and syntactic nuances of language.
\section{Conclusion and Future Work}
\label{sec:conclusion}
We have described SemEval-2024 Task 8 on Multigenerator, Multidomain, and Multilingual Machine-Generated Text Detection. The task garnered significant interest from researchers, with 126, 59, 70, and 30 teams submitting entries for Subtask A Monolingual, Subtask A Multilingual, Subtask B, and Subtask C, respectively. Additionally, we received 54 system description papers before finalizing this submission.

Overall, Subtasks A and B were relatively easier, with all systems showing improvements over the baseline. However, Subtask C proved to be significantly more challenging. Fewer teams participated, and many struggled to surpass our baseline results set in \cite{wang2024m4gt}.

In future work, we plan to extend our focus beyond machine-generated text detection to other modalities such as image, speech, and video detection. Additionally, we intend to develop an open-source demonstration system capable of distinguishing between AI-generated content and human-produced content.

\section*{Limitations}
Despite providing a comprehensive dataset that spans multiple languages, generators, and domains across three distinct tasks in machine-generated text detection, our study encounters several limitations that pave the way for future research.

Firstly, the reliance on textual data without access to white-box information, such as token-level probabilities, confines our detection methods to black-box approaches across all tasks. These methods might exhibit reduced effectiveness and struggle to generalize across new domains, generators, and languages. Additionally, they are susceptible to language-style attacks, including paraphrasing in different tones, back-translation, and other forms of textual adversarial tactics. In contrast, methods that leverage watermarking and white-box patterns show greater promise for robust MGT detection.

Secondly, our approach to boundary identification presupposes that each text comprises an initial segment written by humans followed by machine-generated content, with only one transition point. However, real-world scenarios often present more complex challenges. It is crucial not only to ascertain the presence of mixed text but also to identify all transition points. Texts may originate from human authors and undergo refinement via machine assistance, or vice versa, encompassing machine generation followed by human revision. Addressing these nuanced scenarios will be a focus of our future research efforts.

\section*{Ethics and Broader Impact}
This section outlines potential ethical considerations related to our work.

\paragraph{Data Collection and Licenses}
Our study utilizes pre-existing corpora, specifically the M4 and \outfox datasets, which have been publicly released for research purposes under clear licensing agreements.

\paragraph{Security Implications}
The dataset underpinning our shared task aims to foster the development of robust MGT detection systems. These systems are crucial for identifying and mitigating misuse scenarios, such as curbing the proliferation of automated misinformation campaigns and protecting individuals and institutions from potential financial losses. In fields such as journalism, academia, and legal proceedings, where the authenticity of information is of utmost importance, MGT detection plays a vital role in maintaining content integrity and trust. Furthermore, by enhancing public awareness of the capabilities and limitations of LLMs, we can cultivate a healthy skepticism towards digital content. Effective MGT detection mechanisms are essential for ensuring that users can place their trust in content generated by LLMs.

\section*{Acknowledgments}
We extend our deepest gratitude to the SemEval Shared Task 2024 organizing committee for their enduring patience and support throughout our task's development, and to all participants for their innovative contributions and collaborative spirit during the task coordination phase. Our thanks also go to the anonymous reviewers and program committee chairs, whose constructive feedback has significantly contributed to the improvement of our paper.

\bibliography{custom}
\bibliographystyle{acl_natbib}

\clearpage
\onecolumn
\section*{Appendix}
\appendix

\section{Method Summary}
\label{sec:method-summary}

\subsection{Monolingual Human vs, Machine}
\paragraph{Team Genaios\textsubscript{STA\_mono:1}} \citep{GenaiosSemeval2024task8} achieves the highest accuracy on Subtask A -- Monolingual by 
extracting token-level probabilistic features using four Llama-2 models: Llama-2-7b, Llama-2-7b-chat, Llama-2-13b, and Llama-2-13b-chat. For each token they compute the log probability of the observed token, the log probability of the token predicted by each of the language models, and the entropy of the distribution. These features are then fed to a Transformer Encoder trained in a supervised fashion to detect synthetic text.

\paragraph{Team USTC-BUPT\textsubscript{STA\_mono:2}} \citep{USTC-BUPTSemeval2024task8} incorporates domain adversarial neural networks into the task of machine-generated text detection to reach the second position in the ranking of Subtask A -- Monolingual. They add a gradient reversal layer on top of the baseline, a supervised classifier based on RoBERTa.
In addition, they exploit domain labels to enhance the transferability of learning between training and testing datasets. Their architecture is based on RoBERTa and adds two classification heads, one for category classification (human or synthetic) and one for domain classification (e.g. news, essays, etc.), 
the former uses an MLP layer and the latter is composed of an MLP together with a gradient reversal layer. Finally, the loss is also adapted by summing together the category and the domain losses. The submission evaluated an improvement of approximately 8\% compared to the baseline.

\paragraph{Team PetKaz\textsubscript{STA\_mono:12}} \citep{petkazSemeval2024task8} uses a PLM, RoBERTa-base, fine-tuned for synthetic text detection and enhances it with linguistic features, to train a feed-forward binary classifier (human or synthetic). Their final model uses diverse features and notably, they undersample the human data.

\paragraph{Team HU\textsubscript{STA\_mono:17}} \citep{HUSemeval2024task8} Adopts an architecture trained with a contrastive learning approach based on fine-tuning \emph{sentence-transformers/all-mpnet-base-v2}. The model is trained on an augmented dataset obtained by paraphrasing sentences in the training set.

\paragraph{Team TrustAI\textsubscript{STA\_mono:20}} \citep{TrustAISemeval2024task8} tries two approaches: (a) an ensemble approach with the combination of Multinomial Naive Bayes, LGBM Classifier (lightGBM classifier) and SGD classifier. Each is trained on the concatenation of tf-idf and spaCy embeddings obtained from the Subtask A -- Monolingual dataset and (b) a synthetic text classifier based on RoBERTa fine-tuned first with the outputs of the 1.5B-parameter GPT-2 model and subsequently on the Subtask A -- monolingual dataset. They show that exploring methodologies with different assumptions helps identify the best performing approach.

\paragraph{Team RFBES\textsubscript{STA\_mono:23}} \cite{RFBESSemeval2024task8} 
Both semantic and syntactic considerations were taken into account. For semantic analysis, emphasis was placed on smaller text segments rather than the entire document, operating under the belief that AI models could produce similarly coherent long texts as humans. To achieve this, the XLM-RoBERTa model was employed. Regarding syntactic analysis, a stacked bidirectional LSTM model was used to categorize texts based on their grammatical patterns using UPOS tags. Interestingly, no significant differences in UPOS tag distribution between AI-generated and human-written texts were revealed by the findings.

\paragraph{Team L3i++\textsubscript{STA\_mono:24}} \citep{L3i++Semeval2024task8} Proposes a comparative study among 3 groups of methods to detect synthetic texts: 5 likelihood-based methods; 2 fine-tuned sequence-labeling language models (RoBERTa, XLM-RoBERTa); and a fine-tuned large language model, llama-2-7b. LLaMA 2 outperforms the rest and accurately detects machine-generated texts.

\paragraph{Team art-nat-HHU\textsubscript{STA\_mono:25}} \citep{art-nat-HHU-Semeval2024task8} fine-tunes a RoBERTa model pre-trained for AI-detection and combines it with a set of linguistic features: syntactic, lexical, probabilistic and stylistic. To improve the classifier, they train two separate neural networks on these features, one for each class predicted by the RoBERTa-based classifier.

\paragraph{Team Unibuc - NLP\textsubscript{STA\_mono:28}} \citep{UnibucSemeval2024task8} fine-tunes a Transformer-based model with a MLP as a classification head. They combine the datasets of Subtask A -- monolingual and Subtask B to obtain a larger training set.

\paragraph{NewbieML\textsubscript{STA\_mono:30}} \citep{NewbieMLSemeval2024task8} embeds texts with Longformer-large. Then they Ensemble SVM, LogisticRegression and XGBoost with, as a meta model, a KNN.

\paragraph{Team QUST\textsubscript{STA\_mono:31}} \citep{QUSTSemeval2024task8} experiments with multiple models on a dataset extended through data augmentation. They select the two best-performing models for ensembling: (1) a fine-tuned RoBERTa model, combined with the Multiscale Positive-Unlabeled (MPU) training and (2) a DeBERTa model. They use these two for model fusion through stacking ensemble.

\paragraph{Team NootNoot\textsubscript{STA\_mono:39}} \citep{NootNootSemeval2024task8} carefully fine-tunes a RoBERTa-base model to classify human written and synthetic texts.

\paragraph{Team Mast Kalandar\textsubscript{STA\_mono:40}} \citep{MastKalandarSemeval2024task8} trains a classifier that uses a frozen RoBERTa model with an LSTM head to classify human vs machine written texts.

\paragraph{Team I2C-Huelva\textsubscript{STA\_mono:41}} \citep{I2C-HuelvaSemeval2024task8} proposes a method to use multimodal models together with text analysis to enhance synthetic text detection. To mix the two approaches they explore ensemble by testing several voting methods. 

\paragraph{Team Werkzeug\textsubscript{STA\_mono:45}} \citep{werkzeugSemeval2024task8} uses Roberta-large and XLM-roberta-large to encode texts. To address the anitostropic embedding space created by transformer-based language models, they employ several learnable parametric whitening (PW) transformation. They show that addressing the anisotropicity of the embedding space improves accuracy in detecting synthetic text.

\paragraph{Team NCL-UoR\textsubscript{STA\_mono:50}} \citep{NCLUoRSemeval2024task8} fine-tunes several PLMs including XLM-RoBERTa, RoBERTa with Low-Rank Adaptation (LoRA) and DistilmBERT. Finally, they use majority voting ensembling with XLM-RoBERTa and LoRA-RoBERTa. To confirm that ensembling is a strong technique to boost synthetic text classification accuracy.

\paragraph{Team Sharif-MGTD\textsubscript{STA\_mono:51}} \citep{sharif-MGTD4ADLSemeval2024task8} carefully fine-tunes RoBERTa-base for synthetic text detection, show that pre-trained language models are a versatile approach.

\paragraph{Team BadRock\textsubscript{STA\_mono:*}}~\citep{siino2024badrock} is based on a fine-tuning of a DistilBERT trained on the SST-2 dataset.

\paragraph{Team Collectivized Semantics\textsubscript{STA\_mono:62}} \citep{CollectivizedSemanticsSemeval2024task8} fine-tunes Roberta-base using Ada-LoRa and uses the weighted sum of all the layer hidden states' mean as features to train a classifier. They show that exploiting the knowledge at all layers of encoder language models helps when detecting synthetic texts.

\paragraph{Team IUCL\textsubscript{STA\_mono:63}} \citep{IUStudentTeam} tries both classical ML classifiers, Naive Bayes and Decision Trees as well as fine-tuning transformers and they conclude that fine-tuned RoBERTa is best among the methods they try.

\paragraph{Team SINAI\textsubscript{STA\_mono:67}} \citep{SINAISemeval2024task8} compares three methods: (a) supervised classification, based on fine-tuning the XLM-RoBERTa-Large language model; (b) likelihood-based methods, using GPT-2 to compute the perplexity of each text and use this perplexity as a score; (c) a hybrid approach that merges text with its perplexity value into a classification head. The choice of a mixed approach proves effective in improving synthetic text detection accuracy.

\paragraph{Team MasonTigers\textsubscript{STA\_mono:71}} \citep{MasonTigersSemeval2024task8} experiments with different transformer-based models: Roberta, DistilBERT, ELECTRA and ensembles these models. They also experiment with zero-shot prompting and finetuning FlanT5. Further confirming that ensembling is a strong methodology for detecting synthetic texts.

\paragraph{Team AT\textsubscript{STA\_mono:72}} \citep{TeamATSemeval2024task8} adopts three different semantic embedding algorithms, GLOVE, n-gram embeddings and SentenceBERT as well as their concatenation. The author finds that these pre-trained embeddings, while fast to compute, are not as effective as a fine-tuned RoBERTa model.

\paragraph{Team DUTh\textsubscript{STA\_mono:73}} \citep{DUThSemeval2024task8} experiments with several supervised classification models based on PLMs. Finally, they opt for a fine-tuned mBERT trained for 5 epochs. This approach shows how PLMs fine-tuning is a versatile approach that can be effective when detecting synthetic texts.

\paragraph{Team surbhi\textsubscript{STA\_mono:74}} \citep{SurbhiSemeval2024task8} creates two sets of features (a) stylometric features based on the length of text, the number of words, the average length of words, the number of short words, the proportion of digits and capital letters, individual letters and digits frequencies, hapax-legomena, a measure of text richness, and the frequency of 12 punctuation marks and (b) n-grams: frequencies of the 100 most frequent character-level bi-grams and tri-grams; (c) the output probabilities of fine-tuned Roberta model. Each set of features is used to train a classifier and finally, stylometric and n-gram features are chosen as the best-performing ones. They prove that more classical features can still be valuable when attempting the detection of synthetic text. 

\paragraph{Team Kathlalu\textsubscript{STA\_mono:76}} \citep{KathlaluSemeval2024task8} investigates two methods for constructing a binary classifier to distinguish between human-generated and machine-generated text. The main emphasis is on a straightforward approach based on Zipf's law, which, despite its simplicity, achieves a moderate level of performance. Additionally, they briefly discuss experimentation with the utilization of unigram word counts.

\paragraph{Team KInIT\textsubscript{STA\_mono:77}} \citep{KInITSemeval2024task8} uses two approaches: (a) an ensemble using two-step majority voting for predictions, consisting of 2 LLMs (Falcon-7B and Mistral-7B) fine-tuned using the train set only; (b) 3 zero-shot statistical methods (Entropy, Rank, Binoculars) using Falcon-7B and Falcon-7B-Instruct for calculating the metrics. For classification they use per-language threshold calibration, showing that likelihood-based methods are a viable solution to detect machine-written texts.

\paragraph{Team iimasNLP\textsubscript{STA\_mono:78}} \citep{valdez-EtAl:2024:SemEval2024} fine-tune 4 different language models to identify human and machine generated text, ERNIE, SpanBERT, ConvBERT and XLNet. They find out that RoBERTa is a stronger classifier. In general this shows how fine-tuning PLMs is an effective approach to identify synthetic text.

\paragraph{Team Groningen-F\textsubscript{STA\_mono:86}} \citep{Groningen-FSemeval2024task8} leverage features including tense of the sentence, the voice of the sentence, the sentiment of the sentence, and the number of pronouns vs. proper nouns on the basis of SVM and FFNN models.
The hypothesis here is that traditional models may generalize better than LLMs. It is more computationally effective than LLMs.

\paragraph{Team RUG-D\textsubscript{STA\_mono:100}} \citep{RUG-DSemeval2024task8} fine-tunes different DeBERTa models on a dataset extended with additional synthetic samples. Showing that PLMs fine-tuning is a versatile approach that can be effective in the detection of synthetic texts.  

\paragraph{Team RUG-5\textsubscript{STA\_mono:101}} \citep{RUG-5Semeval2024task8} fine-tunes different pre-trained models for synthetic text classification, distilbert-base-cased for the monolingual tasks and distilbert-base-multilingual-cased for the multilingual ones. Moreover, they explore the use of a Random Forest classifier using frozen distilbert-base-cased embeddings concatenated with 20 linguistic and stylistic features. This approach shows how choosing the right PLMs is crucial for better performance in a given task.


\paragraph{Team Mashee\textsubscript{STA\_mono:115}} \citep{afmzmashee} selects high-quality and low-quality samples using a Chi-square test and adopts the selected samples for few-shot classification using the FlanT5-Large language model. This approach shows how few-shot methodologies can benefit from a careful example selection.

\paragraph{Team TueCICL\textsubscript{STA\_mono:116}} \citep{tueCICLSemeval2024task8} uses a Charachter-level LSTM with pre-trained word2vec embeddings as input to train synthetic text detector. Doing so, they show how one does not necessarily have to use transformers.

\paragraph{Team RUG-1\textsubscript{STA\_mono:117}} \citep{RUG1Semeval2024task8} combines a linear model with document-level features and token-level features that are first passed through an LSTM. Through this methodology, they leverage both local (token-level) and global (document-level) information to identify human-written and synthetic texts.

\paragraph{Team CUNLP\textsubscript{STA\_mono:unknown}}\footnote{Team CUNLP submitted results for development set, but no submissions for the test set, resulting unknown valid rank.} \citep{CUNLPsemeval2024task8} involved employing a range of machine learning techniques, including logistic regression, transformer models, attention mechanisms, and unsupervised learning methods. Through rigorous experimentation, they identified key features influencing classification accuracy, namely text length, vocabulary richness, and coherence. Notably, the highest classification accuracy was achieved by integrating transformer models with TF-IDF representation and feature engineering. However, it is essential to note that this approach demanded substantial computational resources due to the complexity of transformer models and the incorporation of TF-IDF. Additionally, their investigation encompassed a thorough exploration of various ML algorithms, extensive hyperparameter tuning, and optimization techniques. Furthermore, they conducted detailed exploratory data analysis to gain insights into the structural and lexical characteristics of the text data.

\subsection{Multilingual Human vs Machine}
\textbf{Team USTC-BUPT\textsubscript{STA\_Multi:1}} \cite{USTC-BUPTSemeval2024task8}  secured the top position. They initially detect the language of the input text. For English text, they average embeddings from Llama-2-70B, followed by classification through a two-stage CNN. For texts in other languages, the classification problem is transformed into fine-tuning a next-token prediction task using the mT5 model, incorporating special tokens for classification. Their approach integrates both monolingual and multilingual strategies, exploiting large language models for direct embedding extraction and model fine-tuning. This enables the system to adeptly handle text classification across a diverse range of languages, especially those with fewer resources.

\textbf{Team FI Group\textsubscript{STA\_Multi:2}} \cite{FIGroupSemeval2024task8} came in second place. Their methodology began with analyzing latent space distinctions between human and AI-generated texts using Sentence-BERT, hypothesizing that syntax and writing style differences are key. They utilized a hierarchical fusion strategy to adaptively fuse representations from different BERT layers, focusing on syntax over semantics. By classifying each token as Human or AI, their model captures detailed text structures, leveraging the XLM-RoBERTa-Large model for robust multilingual performance.

\textbf{Team KInIT\textsubscript{STA\_Multi:3}} \cite{KInITSemeval2024task8}  placed third by employing an ensemble of two fine-tuned LLMs (Falcon-7B and Mistral-7B) and three zero-shot statistical methods, using a two-step majority voting system. This unique combination of fine-tuned and statistical methods, complemented by language identification and per-language threshold calibration, showcases their innovative approach to integrating diverse techniques for enhanced classification accuracy.

\textbf{Team L3i++\textsubscript{STA\_Multi:5}} \cite{L3i++Semeval2024task8} explored a comparative study among metric-based models, fine-tuned sequence-labeling language models, and a large-scale LLM, finding LLaMA-2 to outperform others in detecting machine-generated texts. Their methodological diversity and comprehensive analysis underline the strengths of fine-tuning LLMs for complex classification tasks across languages.

\textbf{Team QUST\textsubscript{STA\_Multi:6}} \cite{QUSTSemeval2024task8} employed a fine-tuned XLM-RoBERTa model within a stacking ensemble framework, incorporating the MPU framework and DeBERTa model. Their approach emphasizes the efficacy of model fusion and fine-tuning on a multilingual dataset, highlighting the potential of ensemble strategies in enhancing model performance.

\textbf{Team AIpom\textsubscript{STA\_Multi:9}} \cite{AIpomSemEval2024task8} utilized a LoRA-Finetuned LLM for classifying texts as real or fake, achieving notable results with a limited dataset. Their unique approach of using an LLM as a classifier, despite an accidental label swap during training, emphasizes the versatility and potential of LLMs in unconventional scenarios.

\paragraph{Team RFBES\textsubscript{STA\_Multi:10}} \cite{RFBESSemeval2024task8} 
Both semantic and syntactic considerations were taken into account. For semantic analysis, emphasis was placed on smaller text segments rather than the entire document, operating under the belief that AI models could produce similarly coherent long texts as humans. To achieve this, the XLM-RoBERTa model was employed. Regarding syntactic analysis, a stacked bidirectional LSTM model was used to categorize texts based on their grammatical patterns using UPOS tags. Interestingly, no significant differences in UPOS tag distribution between AI-generated and human-written texts were revealed by the findings.

\textbf{Team SINAI\textsubscript{STA\_Multi:21}} \cite{SINAISemeval2024task8}  compared various systems before settling on a fusion model that integrates text with perplexity values for classification. Their comprehensive approach, blending fine-tuning with innovative use of perplexity, offers insightful perspectives on leveraging multiple data dimensions for classification.

\textbf{Team Unibuc-NLP\textsubscript{STA\_Multi:22}} \cite{UnibucSemeval2024task8} focused on exploring different methods of layer selection and fine-tuning within a transformer-based architecture. Their pursuit of optimizing layer interactions for classification tasks highlights the importance of fine-tuning strategies in achieving model effectiveness.

\textbf{Team Werkzeug\textsubscript{STA\_Multi:30}} \cite{werkzeugSemeval2024task8} applied parametric whitening transformations under a mixture-of-experts architecture to address text embedding anisotropy issues. Their methodological innovation, aimed at capturing a broader range of language styles, underscores the potential of advanced architectures in improving classification accuracy.

\textbf{Team RUG-5\textsubscript{STA\_Multi:32}}  \cite{RUG-5Semeval2024task8} augmented DistilBERT with an additional layer for classification, exploring linguistic-stylistic features alongside Random Forest classifiers. Their approach of blending traditional ML techniques with PLMs offers a novel perspective on enhancing text classification through feature integration.

\textbf{Team DUTh\textsubscript{STA\_Multi:33}} \cite{DUThSemeval2024task8}  compared machine learning algorithms and LLMs, ultimately selecting a fine-tuned XLM-RoBERTa model. Their comparative analysis provides valuable insights into the effectiveness of different methodologies for text classification tasks.

\textbf{Team RUG-D\textsubscript{STA\_Multi:39}} \cite{RUG-DSemeval2024task8} used an ensemble of monolingual and multilingual models, testing the performance impact of additional training data. Their ensemble approach and data augmentation strategy highlight the importance of model and data selection in optimizing classification performance.

\textbf{Team MasonTigers\textsubscript{STA\_Multi:49}} \cite{MasonTigersSemeval2024task8} experimented with different transformer models and finetuning strategies, showcasing the effectiveness of ensembling and fine-tuning in addressing classification challenges.

\textbf{Team TrustAI\textsubscript{STA\_Multi:55}} \cite{TrustAISemeval2024task8} focused on fine-tuning the bert-base-multilingual-cased model, demonstrating the potential of pre-trained models in multilingual text classification tasks.

\subsection{Multi-way Detection}
\textbf{Team AISPACE\textsubscript{STB:1}} \cite{AISPACESemeval2024task8} achieves the highest performance in this subtask by fine-tuning various encoder and encoder-decoder models, including RoBERTa, DeBERTa, XLNet, Longformer, and T5. They augment the data with instances from Subtask A and explore the effects of different loss functions and learning rate values. Based on this analysis, they leverage a weighted Cross-Entropy loss to balance samples in different classes. Furthermore, they use an ensemble of different fine-tuned models to improve the robustness of the system. The weights of the models in the ensemble are assigned based on their performance on each class rather than their performance on the whole accuracy.

\textbf{Team Unibuc - NLP\textsubscript{STB:2}} \cite{UnibucSemeval2024task8} use a Transformer-based model with a peculiar two-layer feed-forward network as a classification head. They also augment the data with instances from Subtask A monolingual dataset.

\textbf{Team USTC-BUPT\textsubscript{STB:3}} \cite{USTC-BUPTSemeval2024task8} first leverage the `Llama-2-70B` model to obtain embeddings of the tokens in the text and then average them across all tokens. Next, 
they employ a three-stage classification approach using the CNN classifier.

Firstly, they distinguish between human-generated and machine-generated text using the Llama-2-70B model. Secondly, they categorize ChatGPT and Cohere as a single class for a four-class classification, differentiating them from Davinci, Bloomz, and Dolly. Finally, they perform a binary classification between ChatGPT and Cohere. Despite solid performance, their method does not require fine-tuning.

\textbf{Team L3i++\textsubscript{STB:6}} \cite{L3i++Semeval2024task8} conduct a comparative study among three groups of methods: metric-based models, fine-tuned classification language models (RoBERTa, XLM-R), and a fine-tuned LLM, LLaMA-2-7b. They find LLaMA-2 outperforming the methods from the other groups in MGT detection. The team reveals the analysis of errors and various factors in their paper.

\textbf{Team MLab\textsubscript{STB:7}} \cite{MLabSemeval2024task8} fine-tune DeBERTa and analyze the embeddings from the last layer. They provide insights into the embedding space of the model.

\textbf{Team Werkzeug\textsubscript{STB:8}} \cite{werkzeugSemeval2024task8} utilizes RoBERTa-large and XLM-RoBERTa-large to encode the text. They tackle the problem of anisotropy in text embeddings produced by pre-trained language models (PLMs) by introducing a learnable parametric whitening (PW) transformation. Furthermore, to capture the features of LLM-generated text from different perspectives, they use multiple PW transformation layers as experts under the mixture-of-experts (MoE) architecture equipped with a gating router in their final solution.

\textbf{Team TrustAI\textsubscript{STB:14}} \cite{TrustAISemeval2024task8} explore different pretrained and statistical models for detecting synthetic text, ultimately selecting the RoBERTa-base OpenAI Detector for its effectiveness. This model, originally fine-tuned with outputs from the 1.5B-parameter GPT-2 model, is further fine-tuned on the Subtask-B dataset.

\textbf{Team MGTD4ADL\textsubscript{STB:17}} \cite{MGTD4ADLSemeval2024task8} combine traditional Transformer models (RoBERTa-base, RoBERTa-large, GPT-2-small, XLNet, T5-small) with Sentence Transformers(all-mpnet-base-v2 and all-roberta-large-v1). They further diversify their approach by leveraging different data augmentation techniques and experimenting with various loss functions such as Cross-Entropy (CE), Supervised Contrastive Learning (SCL), and Dual Contrastive Loss (DUALCL).

\textbf{Team scalar\textsubscript{STB:18}} \cite{SCaLARSemeval2024task8} employ an ensemble of three RoBERTa-base models using an individual validation set for each model.

\textbf{Team UMUTeam\textsubscript{23}} \cite{UMUTSemeval2024task8} use fine-tuned RoBERTa model combined with syntactic features of the text such as word length, part of speech, function word frequency, stop-word ratio, and sentence length.

\textbf{Team QUST\textsubscript{STB:36}} \cite{QUSTSemeval2024task8} use fine-tuned RoBERTa and DeBERTa models, integrating them through a stacking ensemble technique.

\textbf{Team MasonTigers\textsubscript{STB:38}} \cite{MasonTigersSemeval2024task8} implement an ensemble of 3 PLMs: RoBERTa, DeBERTa, and ELECTRA. Additionally, they employ zero-shot prompting and use a fine-tuned FLAN-T5 model.

\textbf{Team RUG-5\textsubscript{STB:41}} \cite{RUG-5Semeval2024task8} expands the architecture of DistilBERT models by adding an additional classification layer that incorporates 20 linguistic-stylistic features. They also explore the use of Random Forest classifier on top of embeddings from DistilBERT combined with the same set of linguistic-stylistic features.

\textbf{Team RUG-D\textsubscript{STB:44}} \cite{RUG-DSemeval2024task8} focus on fine-tuning DeBERTa models.

\textbf{Team Groningen-F\textsubscript{STB:45}} \cite{Groningen-FSemeval2024task8} trained traditional machine learning models (SVM and FFNN) with features including tense of the sentence, the voice of the sentence, the sentiment of the sentence, and the number of pronouns vs. proper nouns. 

\textbf{Team DUTh\textsubscript{STB:49}} \cite{DUThSemeval2024task8} explore traditional machine learning algorithms along with BERT for their task. Ultimately, they proceed with BERT fine-tuned for 5 epochs.

\paragraph{Team AT\textsubscript{STB:58}} \citep{TeamATSemeval2024task8} adopts three different semantic embedding algorithms, GLOVE, n-gram embeddings and SentenceBERT as well as their concatenation to identify the generator in Subtask B. The author finds that these pre-trained embeddings, while fast to compute, are not as effective as a fine-tuned RoBERTa model.

\paragraph{Team iimasNLP\textsubscript{STB:61}} \citep{valdez-EtAl:2024:SemEval2024} fine-tune 4 different language models to classify text generated by different models: ERNIE, SpanBERT, ConvBERT and XLNet. They find out that RoBERTa is a stronger classifier. In general this shows how fine-tuning PLMs is an effective approach to identify the generator model.

\textbf{Team CLULab-UofA\textsubscript{STB:62}} \cite{CLULab-UofASemeval2024task8} combine LLM fine-tuning with contrastive learning, specifically using triplet loss.

\subsection{Boundary Identification}

\textbf{Team TM-TREK\textsubscript{STC:1}} \cite{TM-TREKSemEval2024task8} achieved the highest performance in Subtask C by employing an ensemble of models including Longformer, Bigbird, and XLNet for long-text sequence labeling. A simple voting mechanism was used to aggregate the output logits. Their innovative strategy also involved integrating LSTM and CRF layers atop various pre-trained language models (PLMs), along with continuous pretraining, fine-tuning, and utilizing dice loss functions to enhance model performance.

\textbf{Team AIpom\textsubscript{STC:2}} \cite{AIpomSemEval2024task8} introduced a two-stage pipeline that combines outputs from an instruction-tuned, decoder-only model (Mistral-7B-OpenOrca) with two encoder-only sequence taggers. Initially, they trained an instruction-tuned autoregressive model to insert a [BREAK] token into input texts, delineating human-written parts from machine-generated ones. Subsequently, these annotated texts were processed by an encoder-based model for sequence tagging, differentiating human-written tokens (0) from machine-generated tokens (1). An additional encoder trained on a blend of raw and annotated texts further refined sequence tagging. The average change point positions predicted by both encoders served as the final boundary estimation.

\textbf{Team USTC-BUPT\textsubscript{STC:3}} \cite{USTC-BUPTSemeval2024task8} approached the task as a token classification challenge, opting to fine-tune a DeBERTa model enhanced by data augmentation techniques derived from the training set. They reported that DeBERTa-base outperformed other models, and explored the efficacy of sequence labeling (e.g., BIOS) in detecting boundaries within mixed texts. The potential of various layers, including CRF and Dropout, was also examined for their impact on system performance.

\textbf{Team RKadiyala\textsubscript{STC:6}} \cite{RkadiyalaSemeval2024task8} focused on fine-tuning various encoder-based models appended with a Conditional Random Field (CRF) layer, noting that Deberta-V3 yielded the best results on the development set.

\textbf{Team DeepPavlov\textsubscript{STC:7}} \cite{DeepPavlovSemeval2024task8} fine-tuned the Deberta-v3 model using a specially prepared dataset with augmented texts, created by modifying prefixes and suffixes of original texts. They emphasized the importance of augmented data quality in achieving a mean absolute error (MAE) of 15.20903.

\textbf{Team RUG-5\textsubscript{STC:17}} \cite{RUG-5Semeval2024task8} utilized an augmented Longformer model, incorporating extra features into the output state of each token to enrich them with contextual information. This approach aimed at improving token-level classification by leveraging linguistic-stylistic features beyond simple PLM optimization.

\textbf{Team TueCICL\textsubscript{STC:22}} \cite{tueCICLSemeval2024task8} experimented with character-level LSTMs and LSTMs using pretrained Word2Vec embeddings, demonstrating that smaller models could compete with transformer models in the boundary detection task.

\textbf{Team jelarson\textsubscript{STC:25}} \cite{jelarson678Semeval2024task8} explored rule-based methods and linear regression techniques, identifying specific patterns in the training data that could inform better data collection practices, such as ensuring a more randomized and unbiased dataset.

\textbf{Team TueSents\textsubscript{STC:26}} \cite{TueSentsSemeval2024task8} extracted textual features at the sentence level using tools like SpaCy and trained a lightweight BiLSTM model for boundary prediction, achieving an accuracy of 0.7 and MAE of less than 0.5 on the development set.

\textbf{Team MasonTigers\textsubscript{STC:27}} \cite{MasonTigersSemeval2024task8} combined TF-IDF, PPMI, and RoBERTa features with linear regression and Elastic Net, culminating in an ensemble approach based on a weighted development set.

\textbf{Team Unibuc-NLP\textsubscript{STC:28}} \cite{UnibucSemeval2024task8} framed the task as a token classification problem, merging character-level features (extracted via CNN) and word embeddings within a BiLSTM model, further exploring the addition of CRF for enhanced performance.


\end{document}